\documentclass[journal]{IEEEtran}
\usepackage{cite}

\ifCLASSINFOpdf
\else
\fi
\usepackage{amsmath}
\usepackage{amssymb}
\usepackage[pdftex]{graphicx}
\usepackage{algorithm}
\usepackage{algorithmic}
\usepackage{url}
\usepackage{color}

\begin{document}
%

\title{Beyond single receptive field: A receptive field fusion-and-stratification network 
for airborne laser scanning point cloud classification}     

\author{Yongqiang Mao,
        Kaiqiang Chen,
        Wenhui Diao,
        Xian Sun,
        Xiaonan Lu,
        Kun Fu,
        Martin Weinmann
\thanks{This work was supported by the National Natural Science Foundation of China (NSFC) under Grant 62171436. (Corresponding author: Kaiqiang Chen.)

Yongqiang Mao, Wenhui Diao, Xian Sun, Xiaonan Lu, and Kun Fu are with the Aerospace Information
Research Institute, Chinese Academy of Sciences, Beijing
100190, China, the Key Laboratory of Network Information System
Technology (NIST), Aerospace Information Research Institute, Chinese
Academy of Sciences, Beijing 100190, China, the University of Chinese
Academy of Sciences and the School of Electronic, Electrical and
Communication Engineering, University of Chinese Academy of Sciences,
Beijing 100190, China (e-mail: maoyongqiang19@mails.ucas.ac.cn; diaowh@aircas.ac.cn; sunxian@aircas.ac.cn; luxiaonan19@mails.ucas.ac.cn; kunfuiecas@gmail.com).
Kaiqiang Chen is with the Aerospace Information Research Institute,
Chinese Academy of Sciences, Beijing 100190, China and the Key Laboratory
of Network Information System Technology (NIST), Aerospace Information
Research Institute, Chinese Academy of Sciences, Beijing 100190,
China (e-mail: chenkaiqiang14@mails.ucas.ac.cn).
Martin Weinmann is with Institute of Photogrammetry and Remote Sensing, Karlsruhe Institute of Technology, Karlsruhe, Germany (e-mail: martin.weinmann@kit.edu)}
}

%
%

\markboth{Journal of \LaTeX\ Class Files,~Vol.~14, No.~8, August~2015}%
{Shell \MakeLowercase{\textit{et al.}}: Bare Demo of IEEEtran.cls for IEEE Journals}
%



\maketitle



\begin{abstract}
The classification of airborne laser scanning (ALS) point clouds is a critical task of remote sensing and photogrammetry fields. 
Although recent deep learning-based methods have achieved satisfactory performance, they have ignored the unicity of the receptive 
field, which makes the ALS point cloud classification remain challenging for the distinguishment of the areas with complex structures 
and extreme scale variations. In this article, for the objective of configuring multi-receptive field features, we propose a novel 
receptive field fusion-and-stratification network (RFFS-Net). With a novel dilated graph convolution (DGConv) and its extension 
annular dilated convolution (ADConv) as basic building blocks, the receptive field fusion process is implemented with the dilated 
and annular graph fusion (DAGFusion) module, which obtains multi-receptive field feature representation through capturing dilated 
and annular graphs with various receptive regions. The stratification of the receptive fields with point sets of different resolutions 
as the calculation bases is performed with Multi-level Decoders nested in RFFS-Net and driven by the multi-level receptive field 
aggregation loss (MRFALoss) to drive the network to learn in the direction of the supervision labels with different resolutions. 
With receptive field fusion-and-stratification, RFFS-Net is more adaptable to the classification of regions with complex structures 
and extreme scale variations in large-scale ALS point clouds. 
Evaluated on the ISPRS Vaihingen 3D dataset, our RFFS-Net significantly outperforms the baseline (\textit{i.e.} PointConv) approach by 5.3\% on 
mF1 and 5.4\% on mIoU, accomplishing an overall accuracy of 82.1\%, an mF1 of 71.6\%, and an mIoU of 58.2\%. The experiments show 
that our RFFS-Net achieves a new state-of-the-art classification performance on powerline, car, and fence classes. 
Furthermore, experiments on the LASDU dataset and the 2019 IEEE-GRSS Data Fusion Contest dataset show that RFFS-Net achieves a new 
state-of-the-art classification performance. The code is available at {\color{magenta}\url{github.com/WingkeungM/RFFS-Net}}.
\end{abstract}
\begin{IEEEkeywords}
Airborne Laser Scanning, Point cloud, Classification, Deep learning, Dilated graph convolution, Multi-scale receptive fields
\end{IEEEkeywords}
\IEEEpeerreviewmaketitle

\section{Introduction}
\begin{figure}
  \centering
  \includegraphics[scale=.2]{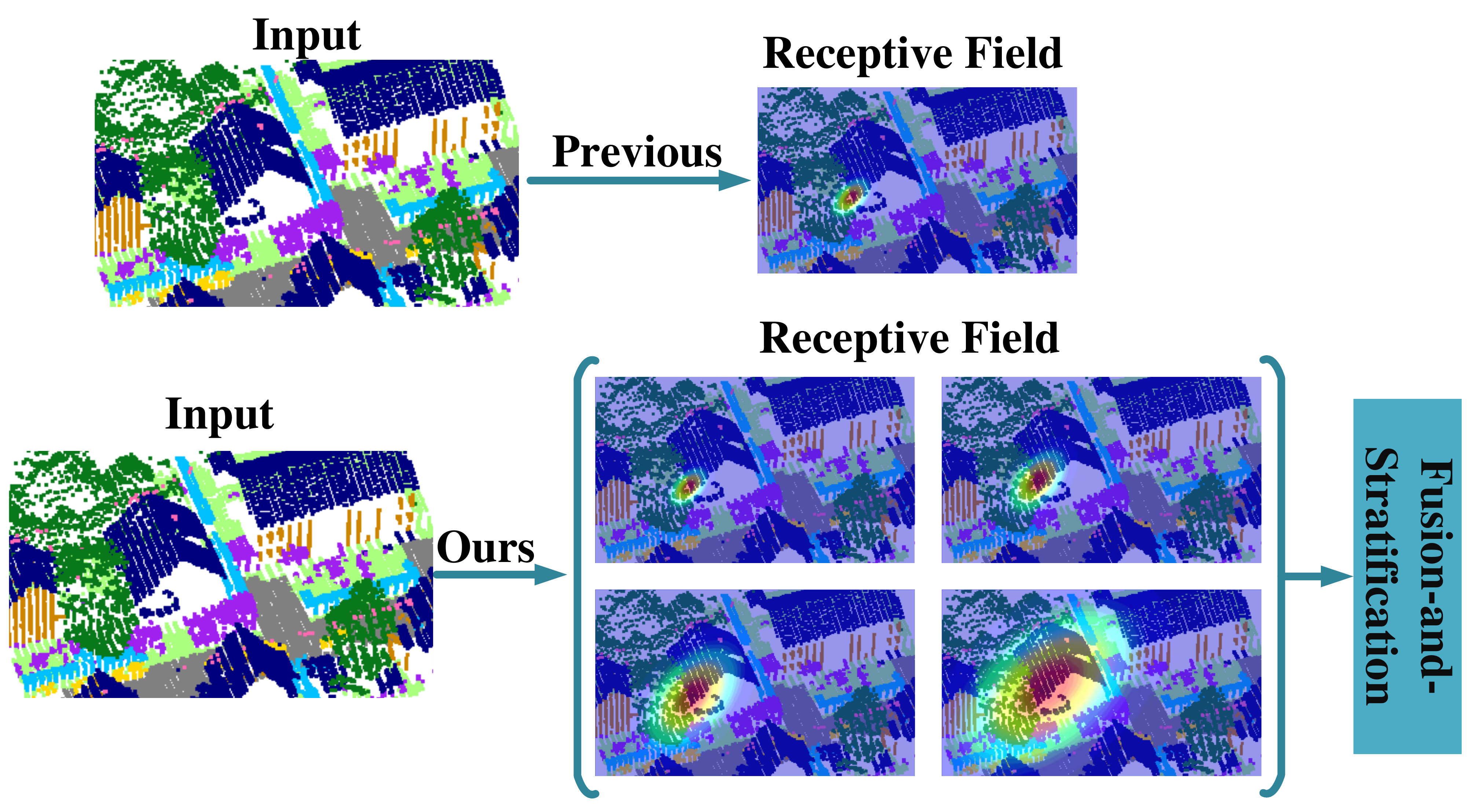}
  \caption{Problem illustration. Sufficient multi-receptive field information is essential for a thorough grasp of areas (Input in the figure) with complex structures and extreme scale variations. The receptive field of previous methods can only perceive local points with less instances and small-scale instances on account of the unicity of receptive fields, while with the fusion-and-stratification of multi-receptive field information, our RFFS-Net can perceive multi-level areas containing various complex structures and instances with various scales.}
\label{FIG:1}
\end{figure}

\IEEEPARstart{W}{ith} the rapid development of 3D data acquisition techniques, such as airborne light detection and ranging (LiDAR) systems, synthetic aperture radar (SAR) systems and dense stereo- or multiview-photogrammetry technology, a large amount of 3D data is meanwhile applied in the fields of remote sensing, photogrammetry, and computer vision. Among all kinds of 3D data, point clouds play an essential role and draw a lot of attention. Point clouds collected by different devices are widely used for a diversity of applications in remote sensing, such as forest monitoring ~\cite{solberg2009mapping,mongus2013computationally}, powerline corridor surveying ~\cite{solberg2009mapping,ene2017large}, 3D building reconstruction ~\cite{kada20093d,bingxun2017variation}, and point cloud classification ~\cite{li2020dance,wen2020directionally}. Due to the demand for 3D data in remote sensing fields, the classification of ALS point clouds has become a crucial task in the fields of remote sensing and photogrammetry. Different from indoor point clouds in the field of computer vision, ALS point clouds suffer from: (1) \textbf{Instances with complex structures.} ALS point cloud scenes usually gather numerous instances with even and uneven structures in an area, such as roof and facade with regular structures, and shrub with irregular structures. (2) \textbf{Instances with extreme scale variations.} The scale of different instances varies greatly, such as the size of a car and a house, which is a problem for many state-of-the-art methods dealing with geospatial data. Therefore, from these two aspects, we infer that the unicity of the receptive field makes the expression of multi-receptive field features remain challenging (Fig. ~\ref{FIG:1}), which causes \textbf{misclassification of regions with complex structures and extreme scale variations}.

In the field of ALS point cloud classification, early researches focus on handcrafted geometric features ~\cite{johnson1999using,weinmann2013visual,Blomley_and_Weinmann_2017}. These methods first extract the local features of point clouds through handcrafted features, and then use machine learning-based classification models, such as Support Vector Machine (SVM) ~\cite{colgan2012mapping} and Random Forest ~\cite{niemeyer2012conditional, niemeyer2014contextual}, to predict the category of each point. However, the extraction of handcrafted features not only requires a cumbersome feature extraction process, but also cannot adapt to the classification task of various scenes with complex structures and extreme scale variations. 

In recent years, with the applications ~\cite{solberg2009mapping,mongus2013computationally} of deep learning methods in remote sensing fields springing up, researchers have also focused on applying deep learning to ALS point cloud classification. Some researchers ~\cite{yang2017convolutional, zhao2018classifying} project raw ALS point clouds into 2D images from different angles, and results of the 2D segmentations then need to be backprojected to 3D space and merged to a consistent labeling with respect to the original point cloud. This kind of projection-based methods brings restrictions on the practical application level to ALS point cloud classification due to the information loss caused by projection onto a regular 2D image grid. Moreover, some researchers ~\cite{li2018pointcnn,thomas2019kpconv} directly use various point convolution operators for feature extraction on raw point clouds. However, these methods do not take into account the insufficiency of deep network receptive fields and the demand for multi-level receptive fields of urban-level ALS point clouds. 

Due to the task for predicting the category point by point, the ALS point cloud classification task requires the network to learn high-level abstract features while retaining more detailed features. Therefore, this makes the network need to have a strong ability to perceive remote information at different distances, that is, the features need to be extracted from multi-scale receptive fields. However, the existing methods are confined to the unicity of the receptive field size and the unicity of the receptive field calculation basis. The universal point cloud classification networks ~\cite{li2020dance, wen2020directionally} employ the fixed KNN search strategy to search for neighboring points, so that the receptive field obtained by using the input point set as the calculation basis is limited to a single region and multi-scale local relationships cannot be captured. Therefore, it causes the network to fail to learn the relationship between points with various structures (such as car and shrub). Besides, the strategy to make the most of multi-level receptive fields is to use a different number of sampling layers, like GADH-Net ~\cite{li2020geometry}. This approach focuses on the receptive field information of different downsampling layers with the point set of original resolution as the calculation basis, ignoring the pivotal role of receptive fields with point sets of different resolutions as the calculation bases. Thus, the unicity of the receptive field size and the unicity of receptive field calculation basis are two thorny issues and will be explained in detail below.

\begin{figure}
  \centering
      \includegraphics[scale=.125]{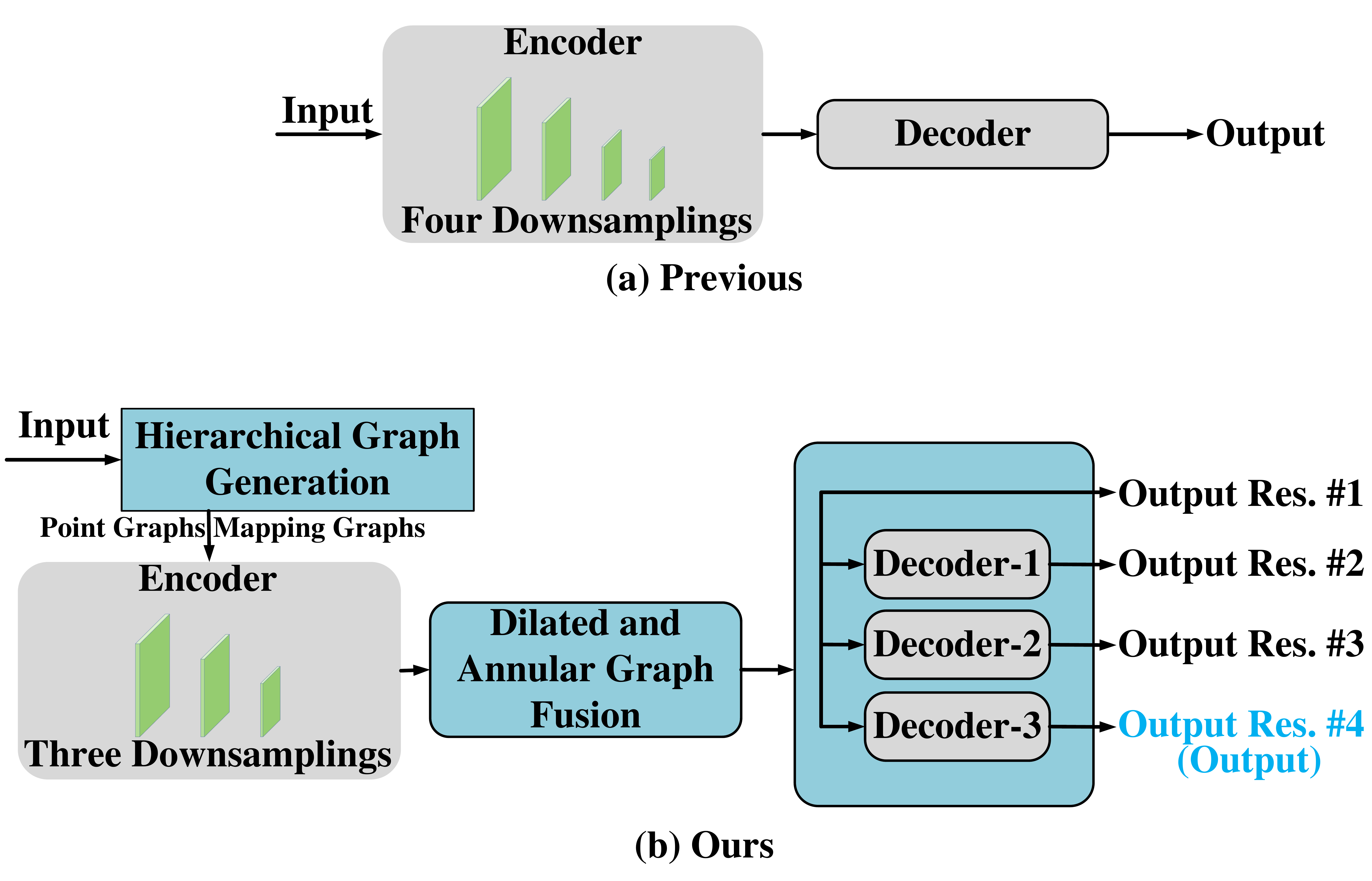}
      \caption{Illustration of the network pipeline. (a) is a previous Encoder-Decoder classification network. The encoder consists four downsampling layers. (b) Our RFFS-Net is a network which uses the point graphs and mapping graphs constructed by the hierarchical graph generation module as the input of the network. The encoder of RFFS-Net only contains three downsampling layers. RFFS-Net uses dilated and annular graph fusion to capture the features with a multi-scale receptive field. Multi-level Decoders are designed for the stratification of receptive fields. 'Output Res. \#1-4' represents the output of different resolutions. 'Output Res. \#4' is the final output. }
\label{FIG:2}
\end{figure}

On the one hand, existing methods are limited to the unicity of the receptive field size. Different from the indoor point cloud data, ALS point clouds are composed of many areas containing complex structures. Most current methods ~\cite{thomas2019kpconv, wang2019graph} extract the point features based on pre-defined neighbors obtained by a fixed KNN graph, which makes it challenging to perceive remote information of areas. Thus, the difficulty in distinguishing areas with complex structures in the ALS point cloud classification caused by the determinism and unicity of the center point's receptive field has never been resolved. In the human visual system, the contrast phenomenon refers to the phenomenon that when two stimuli act on the human eye at the same time, the presence of one stimulus enhances the other stimulus. Since areas with complex structures are spatially dependent, multi-scale receptive fields which capture multiple stimuli (instances) at the same time facilitate the distinction between these stimuli. Inspired by this, we introduce a novel dilated graph convolution (DGConv) to obtain different scales of receptive fields, which captures the dilated graphs of different receptive regions through the proposed Sparse-KNN Search strategy according to the preset dilation rate. 
Further, inspired by the dense connection of DenseNet~\cite{huang2017densely}, we propose a dilated and annular graph fusion (DAGFusion) module taking DGConv and its extension annular dilated convolution (ADConv) as basic building blocks to optimize the expression of multi-receptive field features and solve the misclassification of areas with complex structures.

On the other hand, existing methods are also limited to the unicity of the receptive field calculation basis, which brings challenges to the classification of instances with extreme scale variations in the ALS point cloud scenes. The calculation basis of the existing 3D classification networks' ~\cite{qi2017pointnet++, li2020dance} receptive fields is the input point set, ignoring the pivotal role of point sets with different resolutions as the calculation basis of the receptive field in classifying instances with extreme scale variations. In addition, among existing segmentation frameworks, all recent methods adopt an encoder-decoder structure. U-net~\cite{ronneberger2015u} adopts a fully convolutional approach. In the decoder, upsampling and convolution operations are used for resolution recovery. The decoder of D-LinkNet~\cite{zhou2018d} adopts the transposed-conv operation for feature fusion. RandLA-Net~\cite{hu2020randla} uses up-sampling and MLP to fuse point clouds and restore resolution. However, the output supervision information of these methods is only limited to the label of the same resolution as the input image (or point set), which leads to the low accuracy of the network segmentation of objects with extreme scale variations. Therefore, inspired by the fact that human eyes can distinguish objects according to different references, we regard the point set of each resolution as a calculation basis of the receptive field and introduce Multi-level Decoders that decode the features into the corresponding resolution. Taking Multi-level Decoders as the feature decoding part, we introduce the multi-level receptive field aggregation loss (MRFALoss) which is calculated through the output of each decoder of Multi-level Decoders.

In response to the above two issues and our corresponding solutions, we present a receptive field fusion-and-stratification network (RFFS-Net) for optimizing the expression of multi-receptive field features in the ALS point cloud classification task by receptive field fusion-and-stratification. First, unlike the previous researches ~\cite{li2020dance, qi2017pointnet++,li2018pointcnn} (Fig. \ref{FIG:2}(a)), that directly input the raw point set into the network, we feed point graphs and mapping graphs constructed by a hierarchical graph generation module into the network (Fig. \ref{FIG:2}(b)). Second, considering that consecutive downsampling operations increase the receptive field of the generated features but sacrifice the spatial resolution of the point set, we reduce the number of downsampling layers of the encoder from four to three for preserving the spatial information of the point set (Fig. \ref{FIG:2}). Then, the receptive field fusion process is implemented by the dilated and annular graph fusion (DAGFusion) module (Fig. \ref{FIG:2}(b)) for the fusion of multi-scale receptive field features, which takes the DGConv and its extension ADConv as basic building blocks. DGConv or ADConv can obtain the dilated or annular graphs of different receptive regions through the proposed Sparse-KNN Search strategy according to the preset dilation rate, so as to achieve the purpose of obtaining the features of different receptive fields. DAGFusion merges the features of different receptive fields in the DGConv part and ADConv part respectively through dense connection. The receptive field stratification process defines Multi-level Decoders (Fig. \ref{FIG:2}(b)) nested in RFFS-Net with multi-level receptive field aggregation loss (MRFALoss), which encourages the correct correspondence between the label of each point set with different resolutions and each output of Multi-level Decoders. Regarding the point set of each resolution as the calculation basis for the receptive field, the MRFALoss calculated with each output of Multi-level Decoders and the point label with corresponding resolution penalizes the receptive field information of different scales.

To summarize, our RFFS-Net can capture features with multi-scale receptive fields by receptive field fusion-and-stratification (Fig. \ref{FIG:1}). The code is available at {\color{magenta}\url{github.com/WingkeungM/RFFS-Net}}. The contributions of this study include:
\begin{itemize}
  \item We propose a receptive field fusion-and-stratification network RFFS-Net, achieving the purpose of the expression of multi-receptive field features through receptive field fusion-and-stratification. 
  \item The dilated and annular graph fusion (DAGFusion) module is introduced in this article, which implements receptive field fusion taking the proposed dilated graph convolution (DGConv) and its extension annular dilated convolution (ADConv) as basic building blocks.
  \item We present Multi-level Decoders nested in RFFS-Net driven by the multi-level receptive field aggregation loss (MRFALoss), which not only achieves the stratification process of receptive fields, but also realizes the supervision of the network output by the labels of point sets of different resolutions.
  \item With receptive field fusion-and-stratification, we improve the state-of-the-art of ALS point cloud classification with significant margins on the ISPRS Vaihingen 3D dataset ~\cite{niemeyer2014contextual}, the LASDU dataset ~\cite{ye2020lasdu}, and the 2019 IEEE-GRSS Data Fusion Contest dataset ~\cite{le20192019, bosch2019semantic}. 
\end{itemize}

The remainder of this article is organized as follows: In Section 2, we give a brief review of the researches of airborne laser scanning point cloud classification methods in recent years. In Section 3, the proposed dilated graph convolution (DGConv), dilated and annular graph fusion (DAGFusion) in receptive field fusion process and Multi-level Decoders in receptive field stratification process are introduced. In Section 4, we conduct extensive experiments on the ISPRS Vaihingen 3D dataset ~\cite{niemeyer2014contextual}, the LASDU dataset ~\cite{ye2020lasdu}, and the 2019 IEEE-GRSS Data Fusion Contest dataset ~\cite{le20192019, bosch2019semantic} to show the superior performance of our RFFS-Net. Section 5 gives a comprehensive ablation experiment to verify the effectiveness of our proposed modules. Finally, the paper draws some conclusion remarks in Section 6.

\section{Related Works}
In the following, we will review point cloud classification methods from four aspects: classical machine-learning based methods using handcrafted features (Section 2.1), point feature representation based on deep learning (Section 2.2), dilated convolution on 3D point clouds (Section 2.3), and multi-scale feature expression (Section 2.4).

\subsection{Classical machine-learning based methods using handcrafted features}
In various tasks of point cloud processing, including the classification and segmentation task, it is necessary to extract point cloud geometric features. Classical handcrafted features are modeled by constructing feature descriptors that capture local geometric structures. Feature descriptors are mainly divided into extrinsic descriptors based on spatial coordinates and intrinsic descriptors based on manifolds. Extrinsic descriptors mainly include classical methods such as shape context ~\cite{belongie2000shape}, spin images ~\cite{johnson1999using}, distance-based descriptors ~\cite{ling2007shape}, integral features ~\cite{manay2006integral} and so on. Representative methods of intrinsic descriptors include global point features ~\cite{rustamov2007laplace} and the heat and wave kernel signatures ~\cite{sun2009concise}. After extracting the point features, classical machine learning-based methods build a supervised model for classifying the input point set through classification algorithms, including Support Vector Machine (SVM) ~\cite{colgan2012mapping, garcia2015evolutionary}, Random Forest ~\cite{niemeyer2012conditional, niemeyer2014contextual, Blomley_and_Weinmann_2017}, AdaBoost ~\cite{lodha2007aerial}, etc..

However, non-deep-learning methods require tedious point-by-point extraction of handcrafted features. Additionally, handcrafted features cannot adapt to the segmentation of point clouds in various complex scenes, which in turn reduces the classification performance.

\subsection{Point feature representation based on deep learning}
Based on different 3D data representations, point cloud classification approaches relying on deep learning can be divided into four categories: projection-based methods, voxel-based methods, point-based methods, and graph-based methods.

\subsubsection{Projection-based methods}
The methods based on projection are to convert a 3D object or scene into a sequence of 2D projection images corresponding to different viewpoints and different viewing angles. Considering the successful application of 2D convolutional neural networks, some works ~\cite{su2015multi} project point cloud data onto 2D planes at different angles in 3D space to form 2D images. Then, they use the superior neural network in the image domain to perform feature extraction and follow-up tasks. Specifically, ~\cite{yang2017convolutional} proposes a method to generate an image from an ALS point cloud, which takes the local and global geometric features and full-waveform features of each point in the airborne LiDAR point cloud into 2D feature images. Then, the traditional 2D deep classification network is employed to classify the feature images. Zhao et al. ~\cite{zhao2018classifying} convert ALS point cloud features (such as height, intensity, and roughness) into a set of 32$\times$32 contextual image sequences, and then use convolutional neural networks to classify the point cloud. 

However, these multi-view projection-based methods are sensitive to the choice of the image view angle, and the geometric and spatial information of the data is lost in the process of projection.

\subsubsection{Voxel-based methods}
The existing voxel-based methods ~\cite{ben20183dmfv, meng2019vv} are to convert a three-dimensional object or scene into a common three-dimensional voxel grid. Similarly, the three-dimensional voxel grid can be used for the classic CNN method for feature extraction like a two-dimensional pixel grid. For example, Schmohl and Sörgel ~\cite{schmohl2019submanifold} propose sparse submanifold convolutional networks (SSCNs) to classify voxelized ALS point clouds point by point.

However, the quantization loss of information in the process of obtaining a sparse voxel grid in the voxel-based method depends on the size of the three-dimensional voxel unit. This allows high voxel resolution to retain more information while bringing greater memory and computing power consumption. In contrast, low voxel resolution reduces the memory usage while losing detailed information.

\subsubsection{Point-based methods}
Taking into account the shortcomings of projection-based methods ~\cite{su2015multi, tatarchenko2018tangent} and voxel-based methods ~\cite{le2018pointgrid,tchapmi2017segcloud} in 3D deep learning, PointNet ~\cite{qi2017pointnet} was proposed to directly process point clouds. PointNet independently learns the deep semantic features of each point and then uses the symmetric function (maxpooling) to aggregate the features, which solves the problem of the disorder of points. On this basis, a large number of point-based methods have sprung up. 

\textbf{Point-wise MLP methods}  
~\cite{qi2017pointnet++,jiang2018pointsift} use a shared MLP as the basic unit of the network for feature extraction. ~\cite{yousefhussien2018multi} presents a novel 1D-fully convolutional network that consumes terrain-normalized points directly with the corresponding spectral data to generate pointwise labeling while implicitly learning contextual features in an end-to-end fashion. 

\textbf{Point Convolution methods}  
~\cite{li2018pointcnn,atzmon2018point,thomas2019kpconv,liu2019densepoint} aim to extract high-quality point set features and learn local relationships by designing efficient point convolution operators. ~\cite{wen2020directionally} proposes a multi-scale fully convolutional network with a direction-constrained convolution operator as the basic point feature extraction module. In ~\cite{arief2019addressing}, the Atrous XCRF module is proposed to further optimize the original PointCNN model, and it achieves good performance in the field of ALS point cloud classification. 

However, point-based methods basically focus on the design of efficient point convolution operators, ignoring the unicity of the feature receptive field in the ALS point cloud classification task itself.

\subsubsection{Graph-based methods} 
Unlike point-based methods, ~\cite{wang2019dynamic,wang2019graph} express the point set as a point graph, and use graph convolution to provide an excellent representation of the spatial geometric features of the point set. Through introducing Graph Embedding Module (GEM) and Pyramid Attention Network (PAN), ~\cite{zhiheng2019pyramnet} proposed PyramNet to effectively learn the local feature of point sets. ~{wang2019graph} proposes a Graph Attention Convolution (GAC) to efficiently extract relevant features in local regions by learning the importance of different nodes. SPG~\cite{landrieu2018large} enhances the ability to learn contextual information by constructing a superpoint graphs to extract local features. ~{wen2020airborne} introduces a graph attention convolution module aiming to combine global context information and local structural features and presents a graph attention convolution neural network which can be able to extract the geometric context of ALS point clouds. 

Due to the superior feature extraction ability of graph convolution, we prefer graph-based methods in this paper and introduce dilated and annular graph fusion, and multi-level decoders to extract multi-receptive field features.

\subsection{Dilated convolution on 3D point clouds}
In the image segmentation task, dilated convolution ~{holschneider1990real,giusti2013fast} is proposed to expand the receptive field of features. Compared with the classical convolution, dilated convolution can capture a larger receptive field without increasing the amount of parameters. Similarly, the lack of feature receptive field also exists in the point cloud segmentation task. Dilated convolution in GCNs ~{li2019deepgcns} is proposed through selecting k vertices formed by grabbing one point every other dilation value from the KNN graph. 

Although the dilated convolution has achieved excellent performance in point cloud segmentation of visual fields, it has not yet been involved in the ALS point cloud classification task. Moreover, the vertices obtained by Dilated convolution in GCNs are too sparse, and cannot well express the spatial and geometric relationships of similar points in the graph. In response to this, we propose a dilated graph convolution for ALS point cloud classification that employs a Sparse-KNN search strategy for step sampling to solve the problem of graph sparseness.

\subsection{Multi-scale feature expression}
The irregularity and non-uniformity of the input point sets bring great challenges to the multi-scale expression of features. In particular, PointNet++ ~{qi2017pointnet++} uses a hierarchical structure to extract the local point features of the input point set. From the perspective of multi-scale feature expression, multi-scale grouping and multi-resolution grouping are also introduced for the problems of point cloud caused by non-uniformity and varying density. Later, the Pointwise Pyramid Pooling (3P) module ~{ye20183d} is introduced to capture the coarse-to-fine local feature of multi-scale point sets, and then two-direction hierarchical RNNs are used to further obtain long-range spatial dependence. For airborne LiDAR point clouds, multi-scale feature extraction is used to capture local context information for the center point, and a nonlinear manifold learning method for feature dimension reduction is introduced in ~{huang2020deep}.

Nonetheless, the mentioned approaches still suffer from the limitation of small neighboring regions and cannot capture local-to-global neighbor features. Accompanied by the aim of making full use of the advantages of dilated convolution to obtain multi-scale feature expression, our approach defines a more feasible way by employing DGConv. Inspired by DenseNet's~\cite{huang2017densely} ability to reduce the number of parameters while effectively utilizing features, we apply a dilated and annular graph fusion module and dense connection with DGConv or ADConv as the basic block to point cloud segmentation to obtain multi-scale receptive fields and feature representations.

\section{Method}
In this section, we start with preliminary knowledge of the proposed dilated graph convolution (DGConv) and introduce the proposed dilated and annular graph fusion (DAGFusion) in the receptive field fusion process in Section 3.1. Then, in Section 3.2, we present Multi-level Decoders and multi-level receptive field aggregation loss (MRFALoss) in the receptive field stratification process. Finally, in Section 3.3, the novel network RFFS-Net is given.

\subsection{Receptive field fusion}
Among the point convolution methods, some typical convolutional representative architectures ~\cite{atzmon2018point, thomas2019kpconv} relate kernel weights with fixed kernel points and use a correlation (or interpolation) function to adjust the weight of kernels. Among these methods, KPConv ~\cite{thomas2019kpconv} is well known as a typical structure. However, its hand-crafted fixed kernel composition makes it difficult to handle complex 3D positional changes, and it lacks the ability to adjust the kernel receptive field size. To obtain receptive fields that enable the kernel to obtain variable sizes, we propose a novel dilated graph convolution (DGConv) (Fig. \ref{FIG:3}(a)) to extract features with multi-receptive field.
Further, in order to solve the unicity of the receptive field size, we add a dilated and annular graph fusion (DAGFusion) module with DGConv and its extension annular dilated convolution (ADConv) as the basic blocks, as a multi-receptive field feature extractor.

\subsubsection{Dilated graph convolution}
Considering the graph $\mathcal{G}(\mathcal{V},\mathcal{E})$ constructed by the hierarchical graph generation, $\mathcal{V}=\{1,2,\cdots,N\}$ and $\mathcal{E}\subseteq |\mathcal{V}| \times |\mathcal{V}|$ represent the set of vertices and edges respectively. There exist various ways construct the KNN graph through the one-hop $\mathcal{K}$ nearest neighbors of the center point and thus consider the relationship among the neighboring points in a small region. To get a larger receptive field for the graph, we propose a novel $\mathcal{K}$ nearest neighbor search strategy dubbed sparse $k$-nearest neighbor search (Sparse-KNN Search). 
Our Spaer-KNN Search is based on the original 3D space.
Different from the ordinary $k$-nearest neighbor search, first, our Sparse-KNN Search strategy selects the $\mathcal{K}_{s}$ nearest neighbors in the expansion region according to the dilation rate $r$, which can be formulated as follows:
\begin{equation}
\begin{aligned}
    \mathcal{K}_{s}=&\left \lfloor \frac{\mathcal{K}}{\Delta } \right \rfloor \cdot (r -1 + \Delta) + \\
        &\left \lceil( \frac{\mathcal{K}}{\Delta} - \left \lfloor \frac{\mathcal{K}}{\Delta } \right \rfloor)\cdot (r - 1 + \Delta)  \right \rceil
\end{aligned}
\end{equation}
where $ \left \lfloor \right \rfloor$ is the round down operation, $ \left \lceil \right \rceil$ is the round up operation, and $\Delta$ indicates the sampling step. The calculation process on the right side of the equal sign aims to obtain the smallest expansion region $\mathcal{K}_s$ that can contain  $\mathcal{K}$ nearest neighbors. This is determined by the number of target neighbors $\mathcal{K}$, sampling step $\Delta$ and dilation rate r. Taking into account that the expansion region is an integer, the round-up and round-down operations are used ($\frac{\mathcal{K}}{{\Delta}}$ is not necessarily an integer).  
As claimed in Alg. 1, for each point set $\mathcal{P}=\{p_{i}\in \mathbb{R}^{3}, i=1,2,3, \cdots, N\}$, our Sparse-KNN Search strategy is to sample $\Delta$ points by skipping every $r$ neighbor points based on the selected $\mathcal{K}_{s}$ neighbors.

\begin{figure}
  \centering
      \includegraphics[scale=.18]{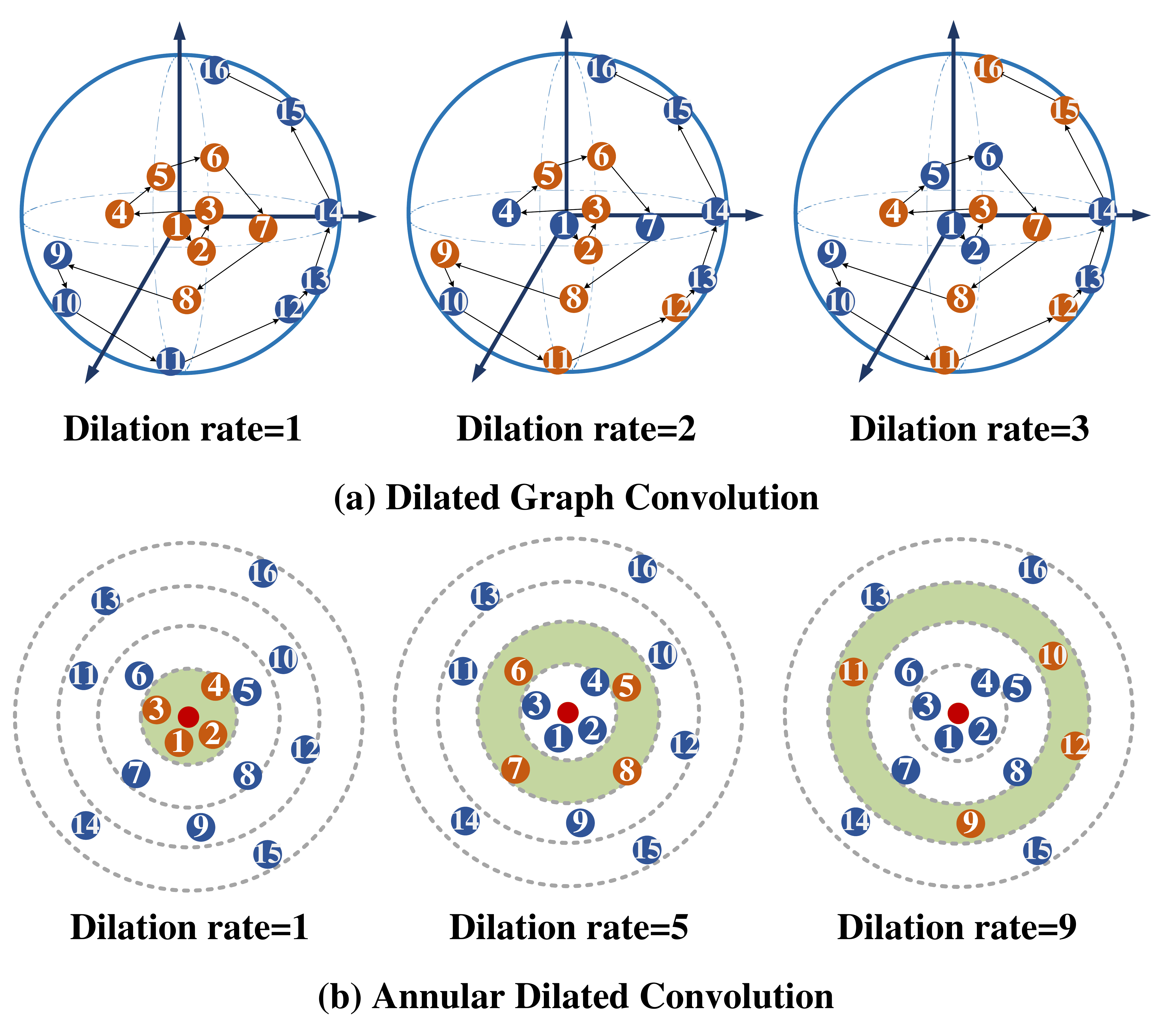}
    \caption{(a): \textbf{The proposed Dilated Graph Convolution.} The diagrams from left to right represent the results of the proposed Sparse-KNN Search strategy with a dilation rate of 1, 2, and 3, respectively. The orange points represent being sampled. The step sampling rate shown is 2. (b): \textbf{Annular Dilated Convolution.} Annular dilated convolutions with a dilation rate of 1, 5, and 9 (left to right). The red point represents the center. The orange points represent being sampled. "Annular" is an abstract ring in the ordering of nearest neighbors.}
  \label{FIG:3}
\end{figure}

After that, we construct a novel dilated graph $\mathcal{G}(\mathcal{V},\mathcal{E}^{r})$, where $\mathcal{E}^{r}$ represents the new edges. 
Because Sparse-KNN Search performs the nearest neighbor search in the 3D space, the dilated graphs is also constructed in the original 3D space.
The new edges $\mathcal{E}^{r}_{i}$ of the vertex $v_{i}$ can be written as:
\begin{equation}
    \begin{aligned}
    \mathcal{E}^{r}_{i} = &\{e_{i,r},\cdots, e_{i,r+\Delta-1}\}\\
    &\cup\{e_{i,2r+\Delta-1}, \cdots, e_{i,2r+2(\Delta-1)}\}\\
    &\cup \cdots \cup \{e_{i,mr+(m-1)(\Delta-1)}, \cdots, e_{i,\mathcal{K}_{s}}\}
    \end{aligned}
\end{equation}

Denote $\mathcal{N}(i)=\{j:(i,j)\ \epsilon\ \mathcal{E}^{r}_{i}\}$ as the neighbor set of vertex $v_{i}$. Our DGConv can be formulated as follows:
\begin{equation}
    \tilde{x_{i}} = \underset{j\in \mathcal{N}(i)}{max} \{\Theta (x_{i}, x_{j})\}
\end{equation}
where $\Theta$ represents the 2D convolution and $max$ is the feature aggregation function. $x_{i}$ indicates the feature descriptor of the central vertex $v_{i}$ and $\{x_{j}, j\in \mathcal{N}(i)\}$ are the features of the neighbor vertex $v_{j}$. 

\textbf{Annular Dilated Convolution (ADConv).} 
In particular, if the sampling value of the nearest neighbor of the center point is equal to the step sampling value ($\bf{\Delta = sample}$), our DGConv becomes a novel annular dilated convolution (ADConv) as depicted in the Fig. \ref{FIG:3}(b). Thus, we can construct the annular graph $\mathcal{G}_{ann}(\mathcal{V},\mathcal{E}_{ann}^{r})$, where $\mathcal{E}_{ann}^{r}$ represents the edges of annular graph. 
It is worth noting that the number of neighbors in the annular graph is $\mathcal{K}_{a}=[(r-1)/\mathcal{K}+1]\cdot \mathcal{K}$. The new edges $\mathcal{E}^{r,i}_{ann}$ of the vertex $v_{i}$ are written as:
\begin{equation}
    \begin{aligned}
    \mathcal{E}^{r,i}_{ann} = &\{e_{i,(n-1)\mathcal{K}+1},\cdots, e_{i,n\mathcal{K}}\}\\
    \end{aligned}
\end{equation}
where $n=(r-1)/\mathcal{K}+1$. Since our ADConv is a ring structure, it is necessary to sample $\mathcal{K}_{a}$ points as the extension region (the smallest region that contains $\mathcal{K}$ neighbors) first, and then sample the $\mathcal{K}$ points at the outermost periphery of the extension region, which is shown in Fig. ~\ref{FIG:3} (b) and Eq. 4.

\begin{algorithm}[tb]
  \caption{Sparse-KNN Search Strategy}
  \label{alg:algorithm}
  \textbf{Input}: N points $\{\bf p_{1}, p_{2}, \cdots, p_{N}\}$\\
  \textbf{Parameter}: number of neighbors $\mathcal{K}_{s}$ after expansion, number of target neighbors $\mathcal{K}$, sampling step $ \Delta $, dilation rate $r$\\ \textbf{Output}: $\{\bf p_{1,\mathcal{K}}, p_{2,\mathcal{K}}, \cdots, p_{N,\mathcal{K}}\}$ N points with $\mathcal{K}$ neighbors
  \begin{algorithmic}[1] 
  \FOR{$\bf p_{m}\in \{\bf p_{1}, p_{2}, \cdots, p_{N} \}$}
  \STATE $p_{m,\mathcal{K}}=\{\}$;
  \STATE {\bf KNN-search: $\bf p_{m,\mathcal{\mathcal{K}}_{s}}\leftarrow p_{m}$};
  \FOR{$i=1\ \textrm{to}\left \lceil \frac{\mathcal{K}_{s}}{r - 1 + \Delta}\right \rceil - 1$}
  \STATE 
  $a=(i-1)  \cdot(r + \Delta - 1) + r$\\
  $b=i \cdot(r + \Delta - 1)$\\
  $p_{m,\mathcal{K}}=p_{m,\mathcal{K}} \cup p_{m,a:b}$;
  \ENDFOR
  \STATE
  $i=\left \lceil \frac{\mathcal{K}_{s}}{r - 1 + \Delta}\right \rceil $\\
  $a= ir + (i-1) \cdot(\Delta -1)$\\
  $b=\mathcal{K}_{s}$\\
  $p_{m,\mathcal{K}}=p_{m,\mathcal{K}} \cup p_{m,a:b}$;
  \ENDFOR
  \STATE \textbf{return} $\forall p_{m,\mathcal{K}} \in \{p_{1,\mathcal{K}}, p_{2,\mathcal{K}}, \cdots, 
  p_{N,\mathcal{K}}\},p_{m,\mathcal{K}} \in \mathbb{R}^{1\times \mathcal{K}} $;
\end{algorithmic}
\end{algorithm}

\begin{figure*}
\centering
    \includegraphics[scale=.085]{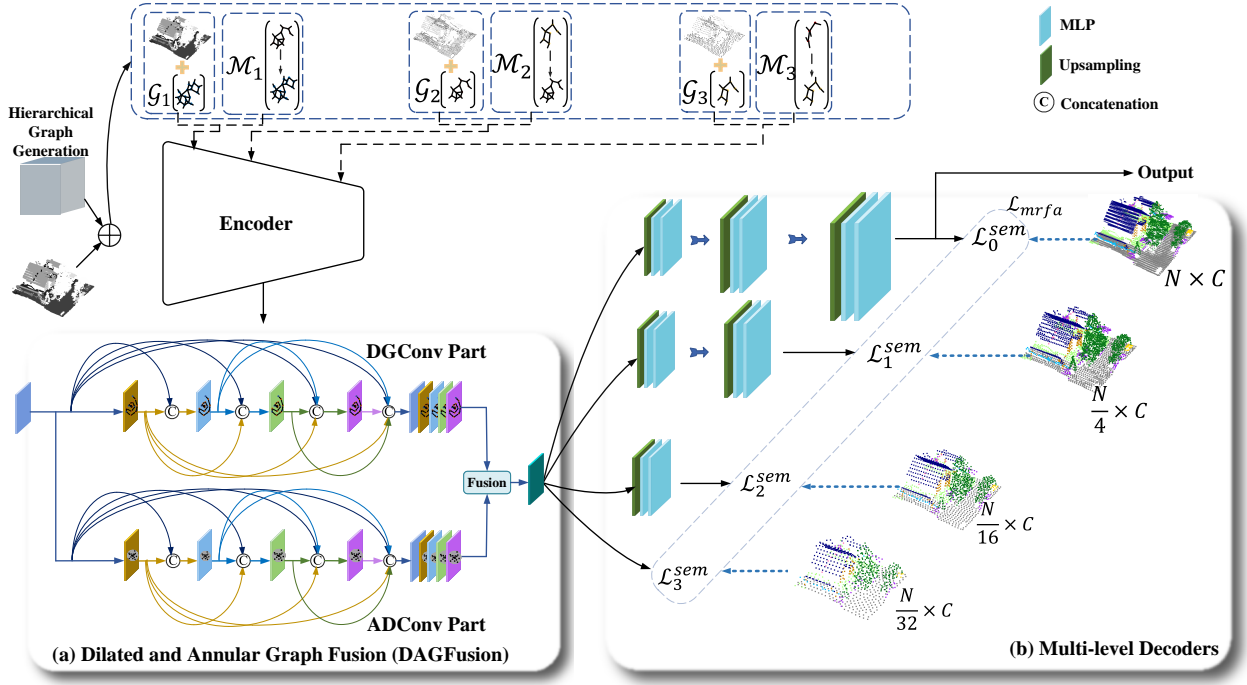}
    \caption{Flowchart of the proposed RFFS-Net. The structure shows the two stages: hierarchical graph generation and encoder-decoder feature extraction and aggregation. (a) represents the proposed DAGFusion module. The multi-level decoders of the network is expressed in (b).}
\label{FIG:4}
\end{figure*}
\subsubsection{Dilated and annular graph fusion}
In order to make better use of DGConv's different dilated graphs with different receptive regions, we add a dilated and annular graph fusion (DAGFusion) module with DGConv and ADConv as the basic blocks as a multi-receptive field feature extractor. In the DAGFusion module, the dense connection ~\cite{huang2017densely} mode is used to fuse convolution outputs with different dilation rates.

The structure of DAGFusion is illustrated in Fig. \ref{FIG:4}(a). DGConv or ADConv operations are organized in a cascade fashion, where the dilation rate of each DGConv or ADConv operation increases layer by layer. Convolutions with small dilation rates are set in the shallower part, while convolutions with larger dilation rates are set in the deeper part. The input of the following operation is the concatenated feature map of all the outputs of each shallower DGConv or ADConv operation and the input feature map. The final output of DAGFusion is a point feature map learned by multi-dilation rate DGConvs or ADConvs. 

We use $\mathcal{G}_{r}(\mathcal{F})$ to term the DGConv. Before concatenating the feature maps, each DGConv operation in DAGFusion can be formulated as follows:
\begin{equation}
  \mathcal{F}^{dgconv}_{m}=\mathcal{G}_{r}\{\mathcal{T}\{\mathcal{F}^{dgconv}_{0}, \mathcal{F}^{dgconv}_{1}, \cdots , \mathcal{F}^{dgconv}_{m-1}\}\}
\end{equation}
where $r$ denotes the dilation rate of the $m$-th DGConv operation, and $\mathcal{F}^{dgconv}_{0}$ denotes the input feature map. $\mathcal{T}\{\mathcal{F}^{dgconv}_{0}, \mathcal{F}^{dgconv}_{1}, \cdots , \mathcal{F}^{dgconv}_{m-1}\}$ means the feature map formed by concatenating the outputs from all previous DGConv operations. 

In the same way, each ADConv in DAGFusion is formulated as follows:
\begin{equation}
  \mathcal{F}^{adconv}_{m}=\mathcal{A}_{r}\{\mathcal{T}\{\mathcal{F}^{adconv}_{0}, \mathcal{F}^{adconv}_{1}, \cdots , \mathcal{F}^{adconv}_{m-1}\}\}
\end{equation}
where $\mathcal{A}_{r}$ represents the $m$-th ADConv operation with dilation rate $r$, and $\mathcal{F}^{adconv}_{0}$ denotes the input feature map. $\mathcal{T}\{\mathcal{F}^{adconv}_{0}, \mathcal{F}^{adconv}_{1}, \cdots , \mathcal{F}^{adconv}_{m-1}\}$ means the feature map formed by concatenating the outputs from all previous ADConv operations. 

Therefore, the output of the DGConv part and the ADConv part in DAGFusion can be computed as follows:
\begin{equation}
  \mathcal{F}^{dgconv}_{out}=\mathcal{M}_{1}\{\mathcal{R}\{\mathcal{F}^{dgconv}_{0}, \mathcal{F}^{dgconv}_{1}, \cdots , \mathcal{F}^{dgconv}_{M}\}\}
\end{equation}
\begin{equation}
  \mathcal{F}^{adconv}_{out}=\mathcal{M}_{2}\{\mathcal{R}\{\mathcal{F}^{adconv}_{0}, \mathcal{F}^{adconv}_{1}, \cdots , \mathcal{F}^{adconv}_{M}\}\}
\end{equation}
where $\mathcal{R}$ indicates the fusion function concatenate operation and $\mathcal{M}_{1}, \mathcal{M}_{2}$ indicate the multilayer perceptrons.

Consequently, we write our DAGFusion as:
\begin{equation}
  \mathcal{F}_{out}= \mathcal{M}\{\mathcal{F}^{dgconv}_{out} || \mathcal{F}^{adconv}_{out}\}
\end{equation}
where $||$ indicates the concatenate operation and $\mathcal{M}$ indicates the multilayer perceptron. 

Theoretically, a larger dilation rate is equivalent to a downsampling operation of the input point set. However, the non-sparseness and versatility of our DGConv and the graph fusion of the DAGFusion module effectively solve the problem of information loss caused by the point cloud sparseness. Furthermore, our DAGFusion brings the network mainly two benefits: a denser feature pyramid and a larger receptive field. 

\subsection{Receptive field stratification}
To achieve the target of receptive field stratification, Multi-level Decoders and the multi-level receptive field aggregation loss (MRFALoss) are introduced.

\subsubsection{Multi-level Decoders}
Among the solutions of decoders with different number of layers, Unet++~\cite{zhou2018unet++} is first proposed. Unet++ uses the nesting of multiple U-nets~\cite{ronneberger2015u}, which makes the number of layers of the encoder and the decoder of each U-net the same. The final output is subject to the label of the same resolution as the input image as supervision information. There is no doubt that such an operation can solve the segmentation problem of objects of different scales. However, for smaller objects, although the purpose of preserving resolution can be achieved through fewer downsampling operations (smaller U-net in Unet++), the network cannot obtain deeper semantic features due to the reduction of downsampling operations, which brings the problem of low segmentation accuracy for the smaller objects. 

To address this problem, we propose a scheme of the sharing encoder with Multi-level Decoders. In the encoder part, we adopt a shared three-layer down-sampling encoder for deep semantic feature extraction. Regarding the point set of each resolution as the calculation basis for the receptive field, we upsample the obtained deepest features to the resolution corresponding to the encoder of each layer, respectively. At this time, the output of each decoder is supervised by labels of different resolutions, and these labels of different resolutions are obtained according to the corresponding indices when the encoder is down-sampling. Unlike Unet++~\cite{zhou2018unet++}, our Multi-level Decoders employ the shared encoder, and the supervision information of the loss function is the labels of point sets of different resolutions. The multi-scale supervision proposed by us acquires deep semantic features while ensuring high-precision segmentation of objects with extreme scale variations.

\begin{figure}
  \centering
    \includegraphics[scale=.09]{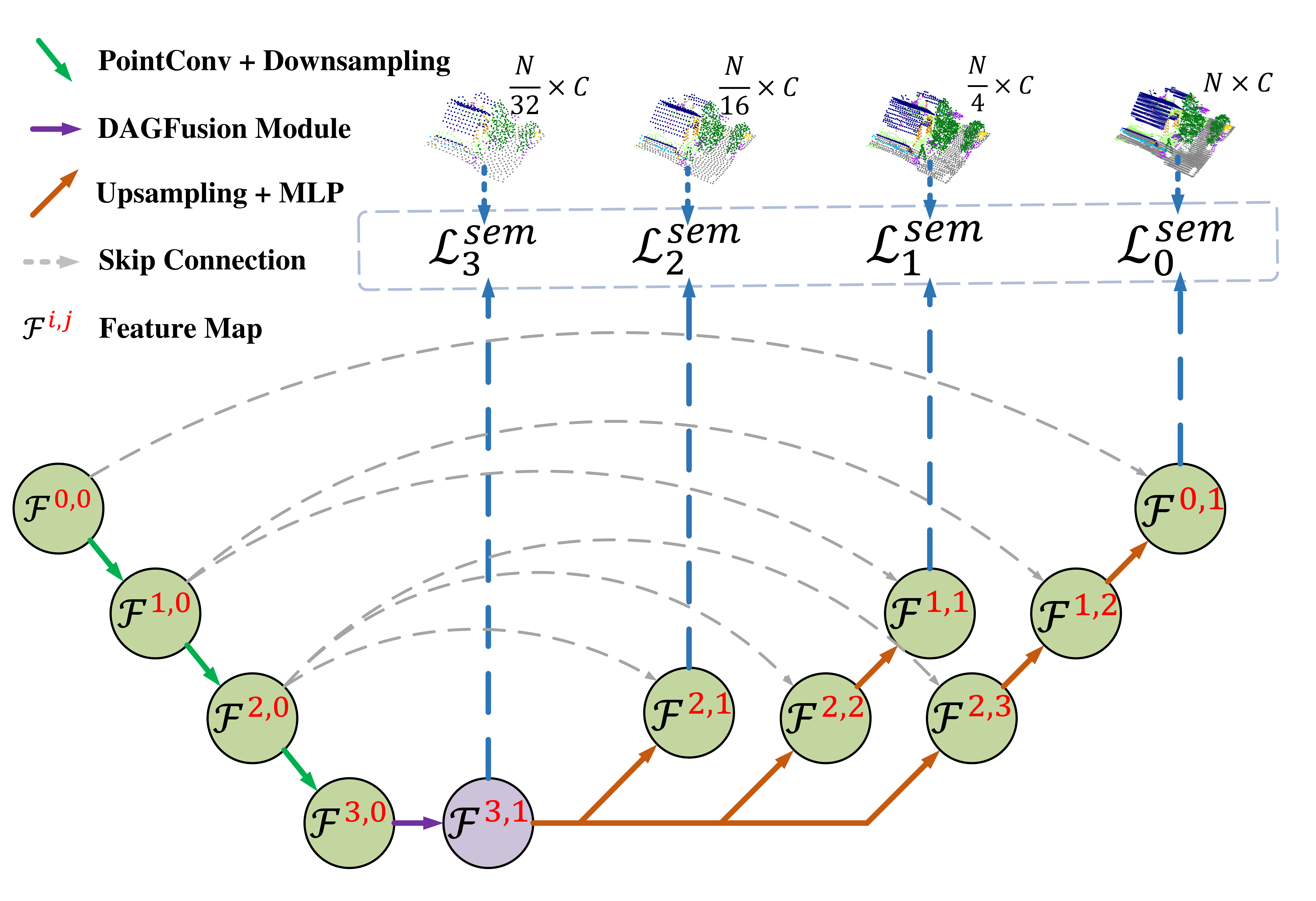}
    \caption{Detailed diagram of RFFS-Net. RFFS-Net consists of a shared encoder, a shared DAGFusion module, and Multi-level Decoders. Specifically, the output of DAGFusion is respectively up-sampled back to different resolution sizes for per-point prediction. The skip pathways are used to connect the encoder and Multi-level Decoders.}
  \label{FIG:5}
\end{figure}

As illustrated in Fig. \ref{FIG:5}, the feature decoding part of RFFS-Net is rooted in Multi-level Decoders with different upsampling layers. Specifically, the output of DAGFusion is up-sampled back to the resolution size of different layers for per-point prediction, respectively. Meanwhile, the skip pathways are used to connect the encoder and Multi-level Decoders. We use $\mathcal{F}^{i,j}$ to represent the feature map, where $i$ indexes the downsampling layers and $j$ indexes different feature maps in the same layer. Thus, each $\mathcal{F}^{i,j}$ can be computed as follows:
\begin{equation}
  \mathcal{F}^{i,j}=\left\{\begin{matrix}
    \Phi(\mathcal{F}^{i-1,j}),&i>0, j=0 \\
  \mathcal{M}(\mathcal{F}^{i,0}\ ||\ \mathcal{F}^{i+1,j+1}), &i<2,j>0\\
  \mathcal{M}(\mathcal{F}^{i,0}\ ||\ \mathcal{F}^{i+1,1}), & i=2,j>0 \\
  DAGFusion(\mathcal{F}_{i,j-1}),&i=3,j=1\\
  \end{matrix}\right.
\end{equation}
where $\Phi$ represents each feature extractor operation (Section 3.3.2) and $\mathcal{M}$ indicates the multilayer perceptron. Besides, $||$ is the concatenate operation.

\subsubsection{Multi-level receptive field aggregation loss}
Before feeding the input point set into the network, labels of different resolutions $\{\mathcal{A}_{0}\in \mathbb{R}^{N\times C}, \mathcal{A}_{1}\in \mathbb{R}^{N/4\times C}, \mathcal{A}_{2}\in \mathbb{R}^{N/16\times C}, \mathcal{A}_{3}\in \mathbb{R}^{N/32\times C}\}$ which are shown at the top of Fig. \ref{FIG:5}, are sampled according to indexes of the farthest point sampling operation. Using the outputs of the Multi-level Decoders, the fully connected layers predict the confidence scores for all candidate semantic categories. Generally, we use the cross-entropy loss $\mathcal{L}^{sem}$ as the semantic segmentation loss. For leveraging multi-level receptive fields, the multi-level receptive field aggregation loss (MRFALoss) $\mathcal{L}_{mrfa}$ is calculated by each output $\{\mathcal{\hat{S}}_{0}, \mathcal{\hat{S}}_{1},\mathcal{\hat{S}}_{2},\mathcal{\hat{S}}_{3}\}$ of Multi-level Decoders and the label with the corresponding resolution. Hence, the loss function $\mathcal{L}_{mrfa}$ is formulated as follows:
\begin{equation}
    \begin{aligned}
        &\mathcal{L}_{mrfa}=\sum_{i=0}^{3}\lambda_{i}\mathcal{L}^{sem}_{i}\\ 
                            &=\sum_{i=0}^{3}\lambda_{i} \sum_{j=1}^{N_{i}}\sum_{c=1}^{C}[\mathcal{A}^{cj}_{i}log \mathcal{\hat{S}}^{cj}_{i} +(1-\mathcal{A}^{cj}_{i})log(1-\mathcal{\hat{S}}^{cj}_{i} )] 
    \end{aligned}
\end{equation}
where $\lambda_{i}$ is a weight hyperparameter for the loss of each Multi-level Decoder and $C$ indicates the number of categories.

\subsection{Receptive field fusion-and-stratification network}
The flowchart illustrated in Fig. \ref{FIG:4} shows the whole architecture of our receptive field fusion-and-stratification network RFFS-Net. The design of our method employs the widely used encoder-decoder architecture. We select the PointConv ~\cite{wu2019pointconv} with three downsampling layers as the baseline. 

As a point cloud classification network, RFFS-Net consists of two stages: hierarchical graph generation and Encoder-Decoders feature extraction and aggregation. The input of our network is given with hierarchical graphs generated through the hierarchical graph generation module in the first stage and features of a raw point set. The second stage is responsible for the representation and aggregation of point cloud features. The two critical modules in the second stage are the dilated and annular graph fusion (DAGFusion) module (Fig. \ref{FIG:4}(a)) and the Multi-level Decoders (Fig. \ref{FIG:4}(b)).

\subsubsection{Hierarchical graph generation}
In the first stage, the raw point set $\mathcal{P}=\{p_{i}\ \in \ \mathbb{R}^{3}, i=1,2,3, \cdots, N\}$ is sent to the hierarchical graph generation module to define the point graphs $\mathcal{G}=\{\mathcal{G}_{1}, \mathcal{G}_{2}, \mathcal{G}_{3}\}$ which is shown above the encoder part in Fig. \ref{FIG:4}. Furthermore, the sampling mapping graphs $\mathcal{M}=\{\mathcal{M}_{1}, \mathcal{M}_{2}, \mathcal{M}_{3}\}$ are constructed at the same time as defining the point graphs. 

The point graphs $\mathcal{G}=\{\mathcal{G}_{1}, \mathcal{G}_{2}, \mathcal{G}_{3}\}$ are constructed by $K$-nearest neighbors, which are sampled with a parameter of 32 in our experiments. Here, each vertex of the point graphs is associated with a point, and the edges of the point graphs are constructed by the $K$-nearest neighbors. The mapping graphs $\mathcal{M}=\{\mathcal{M}_{1}, \mathcal{M}_{2}, \mathcal{M}_{3}\}$ first sample the point set of each layer through the furthest point sampling (FPS) algorithm, and then extract the new graphs from $\mathcal{G}$ according to the index of the FPS algorithm. These sampled graphs are used to form the mapping graphs.

\subsubsection{Encoder-Decoders feature extraction and aggregation}
Following the success of the U-Net-type segmentation networks ~\cite{ronneberger2015u,li2020dance}, we employ the encoder-decoder architecture to build our model. RFFS-Net consists of Multi-level Decoders with different upsampling layers and a common encoder. However, compared with other methods, our RFFS-Net only consists of three downsampling operations in the encoder part. Moreover, the feature decoding part of our network is composed of Multi-level Decoders, which forms our network RFFS-Net. 

Following PointConv ~\cite{wu2019pointconv}, the PointConv operator is used to compose our feature extractor in the encoder part. Therefore, each feature extractor is formulated as follows:
\begin{equation}
    \mathcal{F}_{out} = \Phi(\mathcal{F}_{in}) = PointConv(\mathcal{F}_{in})
\end{equation}
where $\Phi$ denotes each feature extractor operation and $PointConv$ indicates the PointConv operator. $\mathcal{F}_{in}$ and $\mathcal{F}_{out}$ represent the input feature map and output feature map, respectively.

\section{Experiments and results}
In this section, we present the results of conducted experiments to prove the effectiveness of the proposed RFFS-Net for airborne laser scanning point cloud classification. Our experiments are presented according to the following arrangement: In Section 4.1, we introduce the point cloud datasets used in the experiments. The data preprocessing performed on each dataset is described in Section 4.2. Subsequently, the evaluation metrics and the implementation details of our experiments are summarized in Section 4.3 and Section 4.4, respectively. In Section 4.5, Section 4.6, and Section 4.7, the results and visualizations achieved on the ISPRS Vaihingen 3D dataset ~\cite{niemeyer2014contextual}, the LASDU dataset ~\cite{ye2020lasdu}, and the 2019 IEEE-GRSS Data Fusion Contest dataset ~\cite{le20192019, bosch2019semantic} are given, respectively.

\subsection{Experimental datasets}
Extensive experiments are conducted for evaluating the performance of our RFFS-Net on three airborne laser scanning point cloud benchmarks, including the ISPRS 3D Semantic Labeling benchmark dataset (ISPRS Vaihingen 3D) ~\cite{niemeyer2014contextual}, the LASDU dataset ~\cite{ye2020lasdu}, and the dataset of 2019 IEEE-GRSS Data Fusion Contest (DFC2019) ~\cite{le20192019, bosch2019semantic}. 

\subsubsection{ISPRS Vaihingen 3D dataset}
The ISPRS Vaihingen 3D dataset ~\cite{Cramer_2010, Rottensteiner_et_al_2012, niemeyer2014contextual} was collected with a Leica ALS50 system at an average altitude of 500m above Vaihingen, Germany. The point density of the dataset is approximately 6.7 points/m$^{2}$, and the feature of each point is composed of XYZ coordinates, reflectivity, return count information and label. ISPRS Vaihingen 3D is composed of nine categories: powerline, low vegetation, impervious surfaces, car, fence/hedge, roof, facade, shrub, and tree. The dataset contains ALS point clouds of three isolated regions. Following the setting of the ISPRS Benchmark on 3D Semantic Labeling ~\cite{niemeyer2014contextual}, the entire dataset is divided into two parts. 
In total, the training data and test data of the dataset contain 753,876 and 411,722 points, respectively. 

\begin{figure*}
	\centering
	  \includegraphics[width=1.0\textwidth]{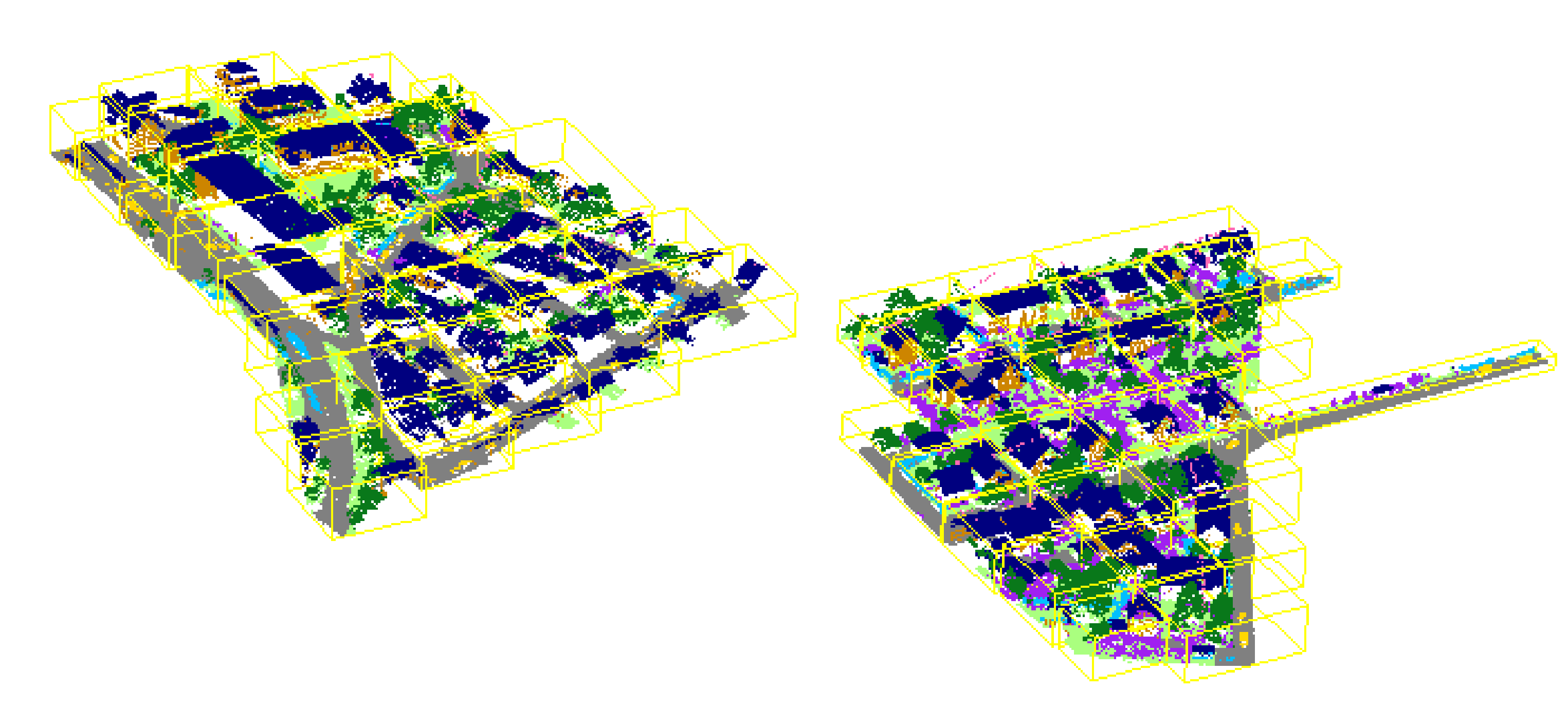}
	\caption{The data preprocessing of the ISPRS Vaihingen 3D dataset. For convenience, the small blocks after dividing on the edge are merged into the surrounding larger blocks.}
	\label{FIG:6}
\end{figure*}

\subsubsection{LASDU dataset}
LASDU ~\cite{ye2020lasdu} is a large-scale aerial LiDAR point cloud dataset acquired with an ALS system of type Leica ALS70 from an altitude of about 1200 m over the valley along the Heihe River in the northwest of China. The dataset covers an urban area with approximately 1 km$^{2}$ of highly-dense residential and industrial buildings, including about 3.12 million points. The average point density is specified as approximately 3-4 pts/$m^2$. The semantic categories of LASDU are composed of Ground, Buildings, Trees, Low Vegetation, and Artifacts. LASDU consists of four regions, namely Section 1, Section 2, Section 3, and Section 4. 

\subsubsection{DFC2019 dataset}
The dataset of 2019 IEEE-GRSS Data Fusion Contest (DFC2019) ~\cite{le20192019, bosch2019semantic} is also an ALS point cloud dataset, which was collected from an area covering about 100 $km^2$ over Jacksonville, Florida and Omaha, Nebraska in the United States. The feature attributes of each point are composed of \{x, y, z, intensity, return number\}. DFC2019 includes 5 categories: Ground, High Vegetation, Buildings, Water, and Bridge Deck.

\subsection{Data preprocessing}
The method based on blocks is adopted to data preprocessing. Here, we only visualize the schematic diagram of blocks on ISPRS Vaihingen 3D.

The original ISPRS Vaihingen 3D dataset covers a large area, so that such huge point cloud data cannot directly be fed into the network. For convenience, we divide each scene of the ISPRS Vaihingen 3D dataset into blocks. Specifically, we divide each scene into regular cuboid blocks (Fig. \ref{FIG:6}) in the horizontal direction. The size of each block is 30m $\times$ 30m within the horizontal dimensions, and there is no area intersection between blocks. When training, we sample 4096 points for each block as the input point set of the network.

For the DFC2019 dataset, before the start of the experiments, we follow D-FCN ~\cite{wen2020directionally} to divide the dataset. The 100 regions of the dataset are selected as the training set, meanwhile the remaining 10 regions of the dataset are selected as the test set. 
Similar to data preprocessing used for the ISPRS Vaihingen 3D dataset, we divide each scene into regular cuboid blocks horizontally, and the size of each area is 75m$\times$75m within the horizontal dimensions. During training, we randomly sample 8192 points as the input point set of the network. 

For the LASDU dataset, we also employ the block division approach for data preprocessing. According to the area size and point density of LASDU, we divide each scene into cuboid blocks covering an area of 50m $\times$ 50m within the horizontal dimensions. Following ~\cite{ye2020lasdu}, we use Sections 2 and 3 of the dataset as the training set for our RFFS-Net, and the remaining Sections 1 and 4 of the dataset as our test set. During training, 4096 points sampled from each scene are fed into the network.

In the test phase, all points in each cuboid block of the three ALS point cloud datasets are directly fed into our RFFS-Net for category prediction.

\subsection{Evaluation metrics}
Like the works on D-FCN ~\cite{wen2020directionally} and DANCE-NET ~\cite{li2020dance}, we focus on the use of the overall accuracy (OA), IoU score, and F1 score to evaluate the performance of our proposed RFFS-Net regarding the task of ALS 3D point cloud classification. Among them, the OA is an index that measures the classification performance of all categories as a whole, and is calculated from the ratio of the correctly classified points to the total number of points, both derived for the test set. The IoU and F1 score also consider the precision and recall of the classification model, and evaluates the performance for each category separately. Therefore, the IoU and F1 score are more suitable for performance evaluation in the case of uneven distribution of categories. The calculation formula for precision and recall is as follows:

\begin{equation}
  precision = \frac{TP}{TP+FP}
\end{equation}
\begin{equation}
  recall = \frac{TP}{TP+FN}
\end{equation}
where TP represents the true positive, FP represents the false positive, and FN represents the false negative. Then, we can write the F1 and IoU score as:
\begin{equation}
  F1\ score = 2 \cdot \frac{precision \cdot recall}{precision + recall}
\end{equation}
\begin{equation}
  IoU\ score = \frac{TP}{TP+FP+FN}
\end{equation}

The mean F1 score (mF1) and mean IoU (mIoU) are the mean of F1 and IoU score for all categories, respectively.

\subsection{Implementation details}
Our model is trained for 500 epochs with batch size 16 and learning rate 0.002 for each dataset on a single GeForce RTX 3090 GPU. Also, we employ the Adam optimizer ~\cite{kingma2017adam} to minimize the overall loss in Eq. 11 with the weight decay of 0.01. Subsequently, the hyper-parameters of weight $\{\lambda_{0}, \lambda_{1}, \lambda_{2}, \lambda_{3}\}$ for the multi-level receptive field aggregation loss $\mathcal{L}_{mrfa}$ are set to $\{1.0, 0.3, 0.3, 0.3\}$ for ISPRS Vaihingen 3D, $\{1.0, 1.5, 1.5, 1.5\}$ for LASDU, and $\{1.0, 1.5, 1.5, 1.5\}$ for DFC2019 through experiments. During the training process, we construct the label graphs based on the input graphs. In addition, unlike all other point cloud classification networks with four downsampling layers, our RFFS-Net only consists of three downsampling layers to form the encoder.
\begin{table*}[htb]
  \caption{\textbf{Classification performance of our RFFS-Net and other models on the ISPRS Vaihingen 3D dataset.} 
  In the first 9 columns, we report the F1 score of each category, meanwhile the overall accuracy (OA), mean F1 score (mF1) and mean IoU score (mIoU) are given in the last three columns.
  }\label{tbl1}
  \centering
  \begin{tabular}{l|ccccccccc|ccc}
            \hline
              Method &powerline&low\_veg & imp\_surf & car & fence & roof & facade &shrub & tree & OA & mF1 & mIoU\\
            \hline
              UM~\cite{horvat2016context} &46.1 &79.0 &89.1 &47.7 &5.2 &92.0 &52.7 &40.9 &77.9 &80.8 &59.0 & 46.7\\ 
              WhuY3~\cite{yang2017convolutional}          &37.1 &81.4 &90.1 &63.4 &23.9 &93.4 &47.5 &39.9 &78.0 &82.3 &61.6 & 49.0\\ 
              LUH~\cite{niemeyer2016hierarchical}            &59.6 &77.5 &91.1 &73.1 &34.0 &\bf{94.2} &56.3 &46.6 &\bf{83.1} &81.6 &68.4 & 55.2\\ 
              BIJ\_W~\cite{wang2018deep}    &13.8 &78.5 &90.5 &56.4 &36.3 &92.2 &53.2 &43.3 &78.4 &81.5 &60.3 & 47.8\\
              RIT\_1~\cite{yousefhussien2018multi}    &37.5 &77.9 &91.5 &73.4 &18.0 &94.0 &49.3 &45.9 &82.5 &81.6 &63.3 & 51.2\\ 
              NANJ2~\cite{zhao2018classifying}        &62.0 &\bf{88.8} &91.2 &66.7 &40.7 &93.6 &42.6 &\bf{55.9} &82.6 &\bf{85.2} &69.3 & 56.5\\ 
              D-FCN~\cite{wen2020directionally}      &70.4 &80.2 &91.4 &78.1 &37.0 &93.0 &\bf{60.5} &46.0 &79.4 &82.2 &70.7 &57.6\\ 
              DANCE-NET~\cite{li2020dance}    &68.4 &81.6 &\bf{92.8} &77.2 &38.6 &93.9 &60.2 &47.2 &81.4 &83.9 &71.2 & \bf{58.3}\\
            \hline
            RFFS-Net    &\bf{75.5} &80.0 &90.5 &\bf{78.5} &\bf{45.5} &92.7 &57.9 &48.3 &75.7 &82.1 &\bf{71.6} & 58.2\\
            \hline
\end{tabular}
\end{table*}
\begin{figure}
  \centering
  \includegraphics[scale=0.44]{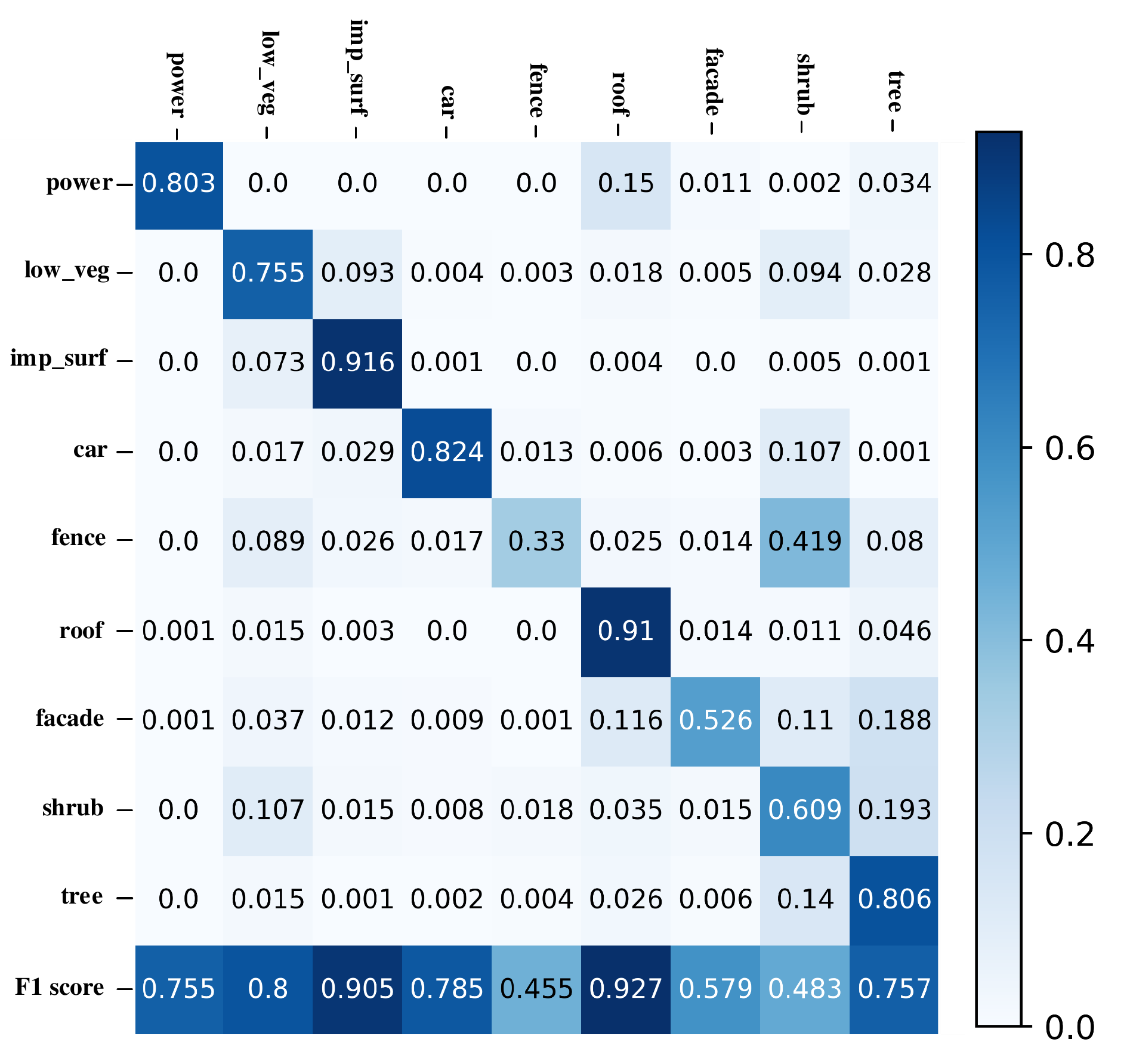}
  \caption{The classification confusion matrix achieved by our RFFS-Net. The F1 score of each category is also given. The scores in the matrix are normalized along each column.}
  \label{FIG:7}
\end{figure}

\begin{figure*}
  \centering
    \includegraphics[width=0.90\textwidth]{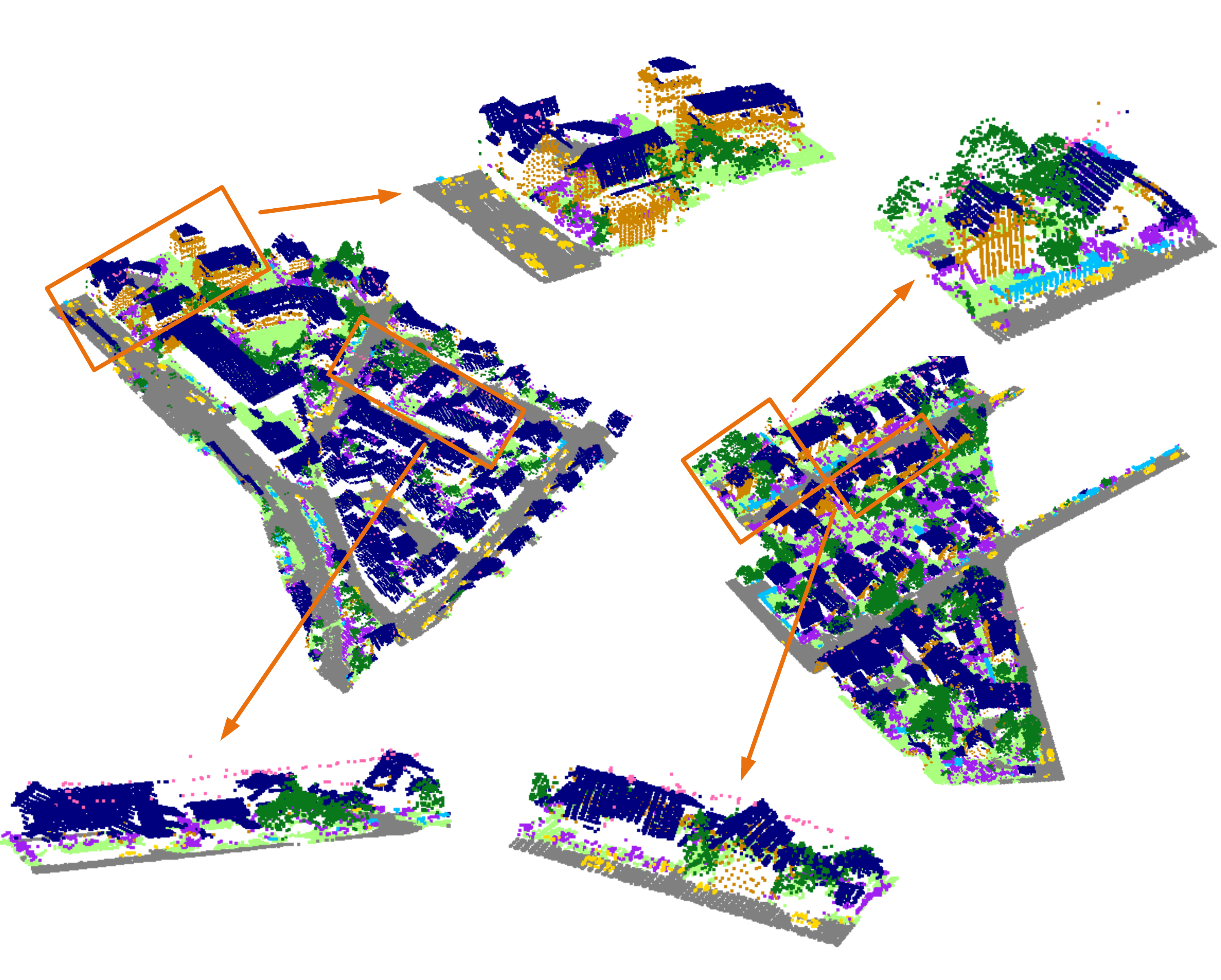}
    \caption{The visualization of the classification results achieved by our RFFS-Net on the ISPRS Vaihingen 3D test set. The enlarged regions show the excellent classification performance of our RFFS-Net for areas containing powerlines (instance with small scale) and complex structures.}
\label{FIG:8}
\end{figure*}

\subsection{Results for the ISPRS Vaihingen 3D dataset}
\subsubsection{Classification results and visualization}
Fig. \ref{FIG:7} lists the confusion matrix corresponding to the classification results achieved by our RFFS-Net. Through the confusion matrix, it is obvious that our RFFS-Net can correctly distinguish all categories of points. Especially for the category of powerline, low vegetation, impervious surfaces, car, roof, and tree, the F1 scores are all above 75\%. 

\begin{figure*}
	\centering
	  \includegraphics[width=1.0\textwidth]{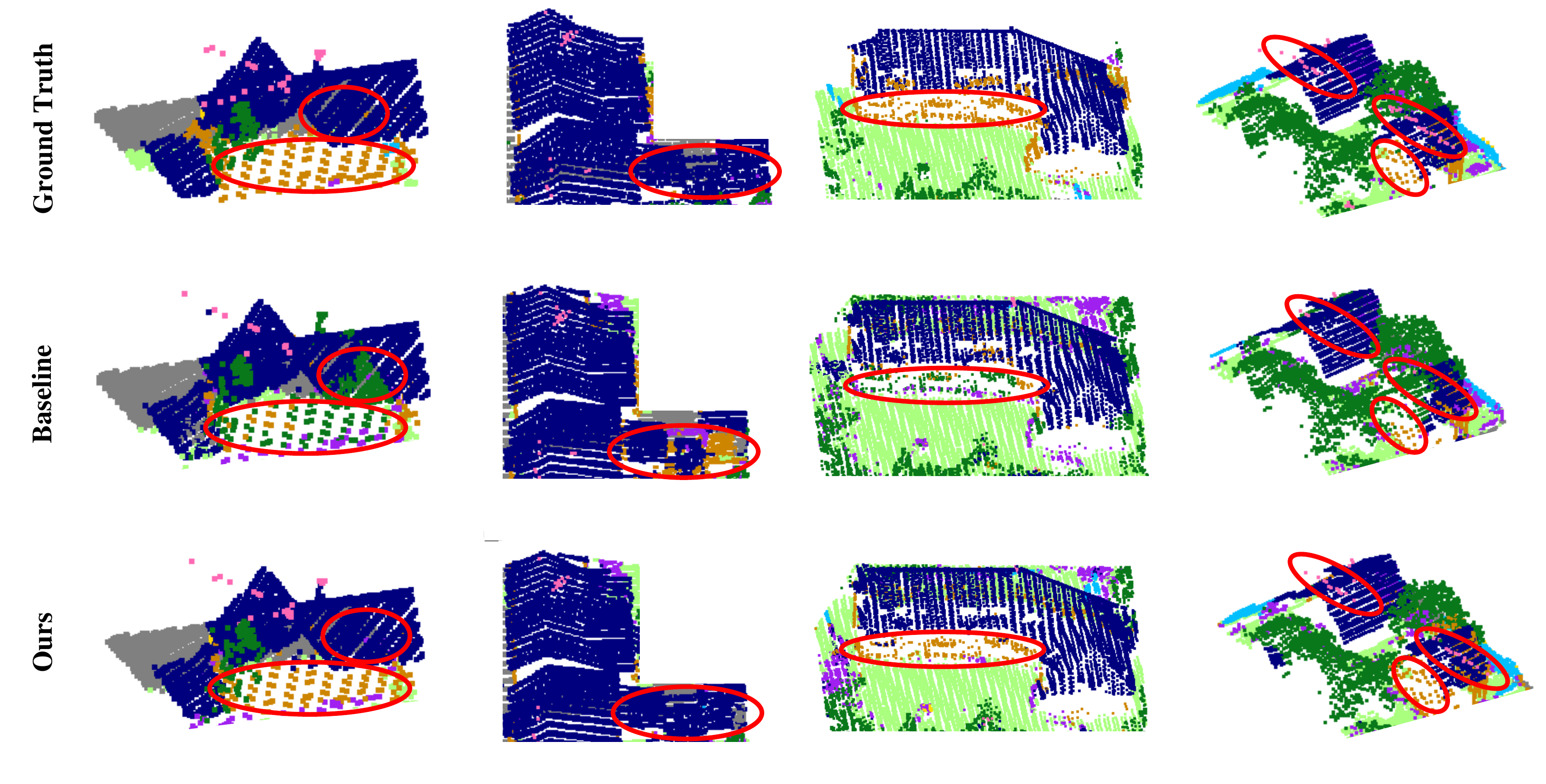}
	\caption{The visual comparison of the ground truth, the classification results achieved by the baseline and our RFFS-Net. The ground truth labels are shown in the first column, the baseline classification results are shown in the second column, and our RFFS-Net results are shown in the last column. A comparison with the baseline reveals that, in specific parts of the scene (shown in the red ellipses), our RFFS-Net can distinguish the given instances better.}
	\label{FIG:9}
\end{figure*}

The visualization of our RFFS-Net prediction results is given in Fig. \ref{FIG:8}. Obviously, our RFFS-Net can correctly classify most of the points. Furthermore, we can observe that the two parts enlarged at the bottom of Fig. \ref{FIG:8} are typical areas that contain powerlines. It is worth noting that the points of semantic categories such as powerline are sparse and has few sampling points. Nevertheless, our proposed RFFS-Net can still distinguish them well. The reason lies in the fact that the encoder of our network is only composed of three downsampling layers, which allows the spatial information of smaller objects and objects with a small number of samples to be well preserved. Not only that, our Multi-level Decoders can well adapt to instances with extreme scale variations, especially for objects with sparse points and smaller scale such as powerline. In addition, the two parts enlarged at the top of Fig. \ref{FIG:8} are typical areas that contain instances with complex structures. The classification results of the regions with complex structures are closely related to the size of the features' receptive field. Since our RFFS-Net aims to obtain multi-scale receptive field features of the central point, it enables the network to learn local features of different scales and even global features. Therefore, we can see that our RFFS-Net distinguishes them effectively for these two regions. Among them, the points belonging to the categories of car, roof, and facade are well identified. Moreover, Fig. \ref{FIG:9} also shows a comparison of the visualizations of the ground truth, classification results achieved with the baseline and our RFFS-Net. According to these visualizations, the classification results of our RFFS-Net are almost consistent with the ground truth. Compared with the baseline, our RFFS-Net has achieved superior classification performance with respect to the categories of facade, roof, powerline, etc.

\begin{table*}[htb]
  \caption{\textbf{Classification performance comparison between our RFFS-Net and other point-based models on the ISPRS Vaihingen 3D dataset.} 
    In the first 9 columns, we report the F1 score of each category, meanwhile the overall accuracy (OA), mean F1 score (mF1) and mean IoU score (mIoU) are given in the last three columns.
    }\label{tbl2}
    \centering
      \begin{tabular}{l|ccccccccc|ccc}
              \hline
              Method&powerline&low\_veg & imp\_surf & car & fence & roof & facade &shrub & tree & OA & mF1 & mIoU \\
              \hline
              PointNet++~\cite{li2020dance}         &57.9 &79.6 &90.6 &66.1 &31.5 &91.6 &54.3 &41.6 &77.0 &81.2 &65.6 & 52.0\\ 
              PointSIFT~\cite{li2020dance}         &55.7 &80.7 &90.9 &77.8 &30.5 &92.5 &56.9 &44.4 &79.6 &82.2 &67.7 & 54.6\\
              PointCNN~\cite{li2020dance}          &61.5 &82.7 &91.8 &75.8 &35.9 &92.7 &57.8 &49.1 &78.1 &83.3 &69.5 & 56.3\\ 
              PointCNN + A-XCRF~\cite{li2020dance}  &63.0 &82.6 &\bf{91.9} &74.9 &39.9 &\bf{94.5} &59.3 &\bf{50.8} &\bf{82.7} &\bf{85.0} &71.1 & 58.0\\ 
              KPConv~\cite{li2020dance}            &63.1 &82.3 &91.4 &72.5 &25.2 &94.4 &60.3 &44.9 &81.2 &83.7 &68.4 & 55.7\\
              DGCNN~\cite{wang2019dynamic}            &44.6 &71.2 &81.8 &42.0 &11.8 &93.8 &\bf{64.3} &46.4 &81.7 &78.3 &59.7 & 46.8\\
              RandLA-Net~\cite{huang2021granet}        &68.8 &82.1 &91.3 &76.6 &43.8 &91.1 &61.9 &45.2 &77.4 &82.1 &70.9 & 57.4\\
              GA-Net~\cite{deng2021ga}        &65.6 &\bf{83.3} &90.6 &77.1 &41.6 &93.4 &61.1 &46.9 &80.3 &82.9 &71.1 & 57.9\\
              SCF-Net~\cite{fan2021scf}        &64.2 &81.5 &90.8 &73.9 &35.2 &93.6 &61.5 &43.4 &82.6 &83.2 &69.8 & 56.8\\
              PointConv~\cite{wu2019pointconv}        &65.5 &79.9 &88.5 &72.1 &25.0 &90.5 &54.2 &45.6 &75.8 &79.6 &66.3 & 52.8\\
            \hline
              RFFS-Net    &\bf{75.5} &80.0 &90.5 &\bf{78.5} &\bf{45.5} &92.7 &57.9 &48.3 &75.7 &82.1 &\bf{71.6} & \bf{58.2}\\
              \hline
\end{tabular}
\end{table*}

\subsubsection{Classification performance}
We compared the classification performance achieved by our RFFS-Net with other methods on ISPRS Vaihingen 3D, including UM ~\cite{horvat2016context}, WhuY3 ~\cite{yang2017convolutional}, LUH ~\cite{niemeyer2016hierarchical}, BIJ\_W ~\cite{wang2018deep}, RIT\_1 ~\cite{yousefhussien2018multi}, NANJ2 ~\cite{zhao2018classifying}, D-FCN ~\cite{wen2020directionally}, and DANCE-NET ~\cite{li2020dance}. The comparison of the classification performance between our RFFS-Net and other methods is depicted in Table \ref{tbl1}. Obviously, our method achieves superior classification performance than all other methods, and it reaches the highest mF1. In particular, our RFFS-Net outperforms the state-of-the-art performance of DANCE-NET by 0.4\% on mF1.

In addition, our method achieves the highest classification performance in categories such as powerline and car. First, powerline is an instance where the points are sparse and have few samples. The reason why such instances can be distinguished better is that our Multi-level Decoders in RFFS-Net can make the most of the receptive fields of multiple bases to better adapt to instances with extreme scale variations and sparse sample points. Second, the car is a small-scale instance. Not only that, compared with the scale of the entire scene, the spatial coordinates of cars are similar to the impervious surfaces (such as roads). Under the premise of inputting only xyz coordinates, the category car can easily be misclassified as impervious surfaces. However, our RFFS-Net obtains the information of the different neighborhoods of the central point through DAGFusion, so that the neighborhood of the central point can cover the information of points such as cars and impervious surfaces at the same time. Therefore, our method captures local features of different scales, and achieves the purpose of classifying instances with similar spatial locations. Third, the region where powerline exists also belongs to scenes with complex structures. Such scenes have higher requirements for the network's ability to capture receptive fields of different scales. Therefore, our RFFS-Net can classify the regions with complex structures well.

Moreover, Table \ref{tbl2} lists the performance comparison between our RFFS-Net and several point-based methods (including PointNet++ ~\cite{qi2017pointnet++} and PointSIFT ~\cite{jiang2018pointsift}) recently proposed in the fields of remote sensing and computer vision. Considering the large gap in the number of points for each category contained in ISPRS Vaihingen 3D, focusing on the overall accuracy (OA) may cause categories with a small sample size to be ignored, so we focus on the mF1 and mIoU scores in this paper. Although the encoder of our method has only three downsampling layers, our RFFS-Net has improved mF1 to 71.6\% and mIoU to 58.2\%. It can be seen from Table \ref{tbl2} that our proposed RFFS-Net outperforms all point-based models (including recent RandLA-Net, GA-Net and SCF-Net) on mF1 and mIoU even when compared with PointCNN with the A-XCRF model~\cite{arief2019addressing}. However, PointCNN+A-XCRF takes further post-processing steps to refine the classification results, while our RFFS-Net does not involve any post-processing techniques.

\begin{figure*}
  \centering
    \includegraphics[width=0.95\textwidth]{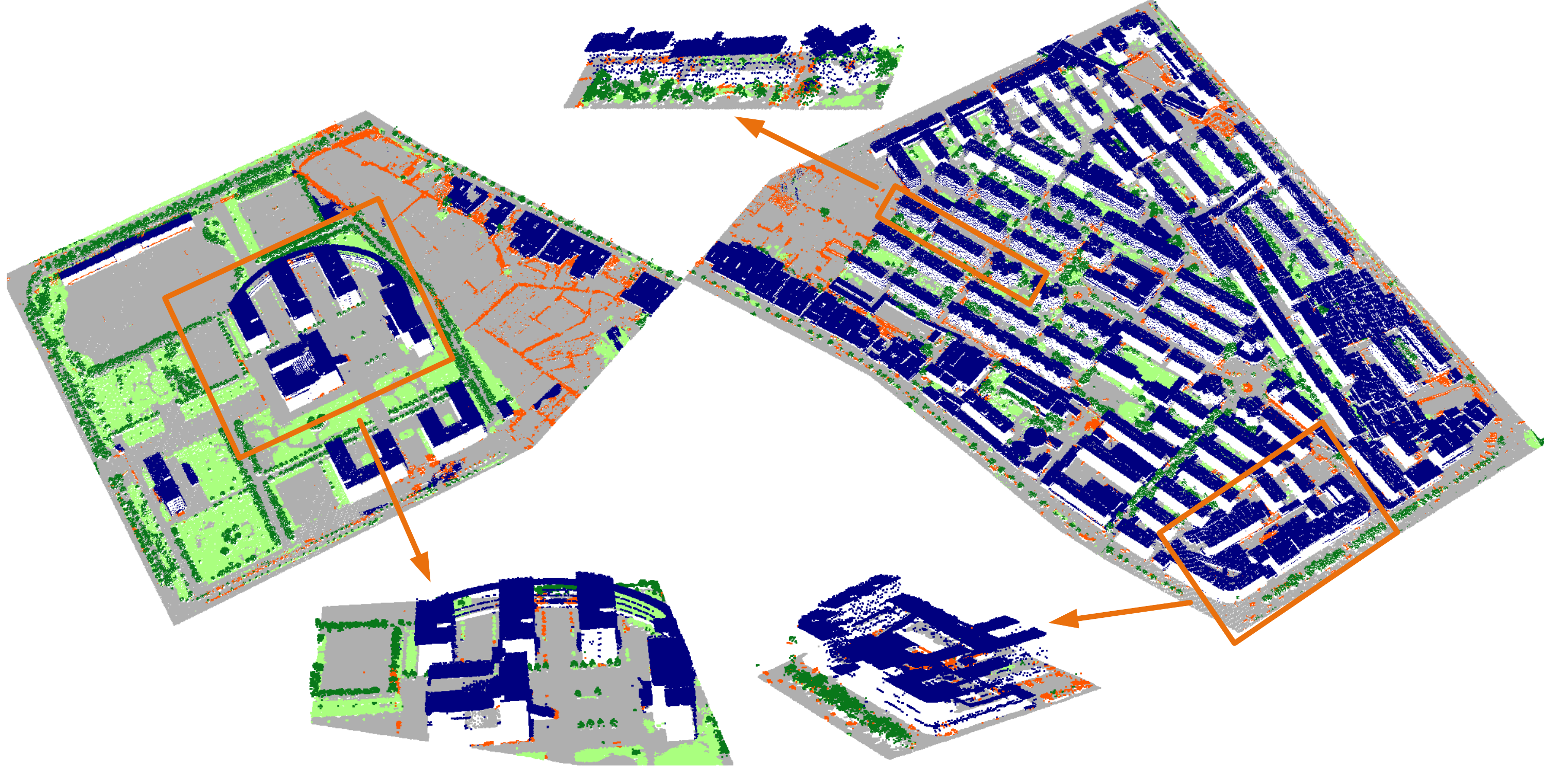}
  \caption{The classification results achieved by our RFFS-Net on the LASDU dataset. The enlarged regions show the excellent classification performance of our RFFS-Net for areas with complex instances.}
  \label{FIG:10}
\end{figure*}

\begin{table*}[htb]
  \caption{\textbf{Comparison of the classification performance between our RFFS-Net and other methods on the LASDU dataset.} 
  In the first 5 columns, we report the F1 score of each category, meanwhile the overall accuracy (OA), mean F1 score (mF1) and mean IoU score (mIoU) 
  are given in the last three columns.
  }\label{tbl3}
  \centering
  \begin{tabular}{c|ccccc|ccc}
  \hline
  Method& Ground & Buildings & Trees & Low vegetation & Artifacts &OA &mF1  & mIoU \\
  \hline
  PointNet++~\cite{ye2020lasdu}         	&87.74 	&90.63 	&81.98 	&63.17 	&31.26 &82.84     &70.96 &59.04 \\
  PointCNN~\cite{li2022vd}  	       	&89.30 	&92.83 	&84.08 	&62.77 	&31.65 &85.04   	&72.13 &60.87\\
  DensePoint~\cite{li2022vd}	     	&89.78 	&94.77 	&85.20 	&65.45 	&34.17  &86.31 	 	&73.87  & 63.00\\
  DGCNN~\cite{wang2019dynamic}	        	&90.52 	&93.21 	&81.55 	&63.26 	&37.08  &85.51 	 	&73.12 & 61.57 \\
  KPConv~\cite{li2022vd}	  	       	&89.12 	&93.43 	&83.22 	&59.70 	&31.85 &83.71 		&71.47 & 60.16\\
  PosPool~\cite{li2022vd}  	         	&88.25 	&93.67 	&83.92 	&61.00 	&38.34 &83.52 		&73.03 & 61.39\\
  HDA-PointNet++~\cite{ye2020lasdu}      	&88.74 	&93.16 	&82.24 	&65.24 	&36.89 &84.37 	  &73.25  & 61.56\\
  PointConv~\cite{wu2019pointconv}  	         	&89.57 	&94.31 	&84.59 	&67.51 	&36.41 &85.91 		&74.48 &63.37 \\
  \hline
  RFFS-Net          &\bf{90.92} &\bf{95.35} &\bf{86.81} &\bf{71.01}&\bf{44.36}&\bf{87.12}&\bf{77.69} & \bf{66.94}\\
  \hline
\end{tabular}
\end{table*}

\subsection{Results for the LASDU dataset}
Table \ref{tbl3} lists the results achieved by our RFFS-Net and other methods on LASDU, including PointNet++ ~\cite{qi2017pointnet++}, PointCNN ~\cite{li2018pointcnn}, DensePoint ~\cite{liu2019densepoint}, DGCNN ~\cite{wang2019dynamic}, KPConv ~\cite{thomas2019kpconv}, PosPool ~\cite{huang2020deep}, PointConv ~\cite{wu2019pointconv} and HDA-PointNet++ ~\cite{liu2020closer}. Obviously, our RFFS-Net achieves the best classification performance, and exceeds the state-of-the-art performance of PointConv by 3.21\%, reaching 77.69\% mF1. It is worth noting that our RFFS-Net achieves the best classification results in all categories, which strongly proves the high performance of our method.

Moreover, the visualization of prediction results achieved for the LASDU dataset is shown in Fig. \ref{FIG:10}. We can find that most points can be distinguished correctly. For better observation, we zoom in three local regions, all of which contain complex structures. As shown by the visualization results of these local regions, it once again proves the effectiveness of our RFFS-Net for classifying complex scenes with abundant structures. The reason lies in the fact that our method can perceive different regions of the central points to extract the features of multi-scale receptive fields.
\begin{table*}[htb]
  \caption{\textbf{Comparison of the classification performance between our RFFS-Net and other methods on the DFC2019 dataset.} 
  In the first 5 columns, we report the F1 score of each category, meanwhile the overall accuracy (OA), mean F1 score (mF1) and mean IoU score 
  (mIoU) are given in the last three columns.
  }\label{tbl4}
  \centering
  \begin{tabular}{l|ccccc|ccc}
    \hline
    Method& Ground & Trees & Buildings & Water & Bridge &OA &mF1   & mIoU\\
    \hline
    PointNet++~\cite{li2020dance}      	&98.30 	&95.80 	&79.70 	&4.40 	&7.30   &92.70 	&57.10 & 52.18 \\
    PointSIFT~\cite{li2020dance}       	&98.60 	&97.00 	&85.50 	&46.40 	&60.40  &94.00   &77.60 & 67.91\\
    PointCNN~\cite{li2020dance}        	&98.70 	&97.20 	&84.90 	&44.10 	&65.30   &93.80   &78.00 & 68.50\\
    KPConv~\cite{li2020dance}           	&98.40 	&94.20 	&87.40 	&43.00 	&77.50  &94.50   &80.10 & 70.83\\
    DGCNN~\cite{wang2019dynamic}         	&97.88 	&93.22 	&90.37 	&\bf{88.23} 	&54.39  &95.08   &84.82 & 76.37\\
    D-FCN~\cite{wen2020directionally}	                   	&\bf{99.10} 	&\bf{98.10} 	&89.90 	&45.00 	&73.00  &95.60   &81.00 &72.53\\
    DANCE-NET~\cite{li2020dance}	        	&\bf{99.10} 	&93.90 	&87.00 	&58.30 	&\bf{83.90}  &\bf{96.80}  &84.40 &75.42\\
    PointConv~\cite{wu2019pointconv}  &97.33 	&95.82 	&\bf{93.63}&74.50 	&69.24  &95.32   &86.10 & 77.42\\\hline
    RFFS-Net          &96.61 &96.10 &88.69 &77.84 &80.97 &94.31   &\bf{88.04} &\bf{79.47}\\
  \hline
\end{tabular}
\end{table*}
\begin{figure*}
  \centering
    \includegraphics[width=\textwidth]{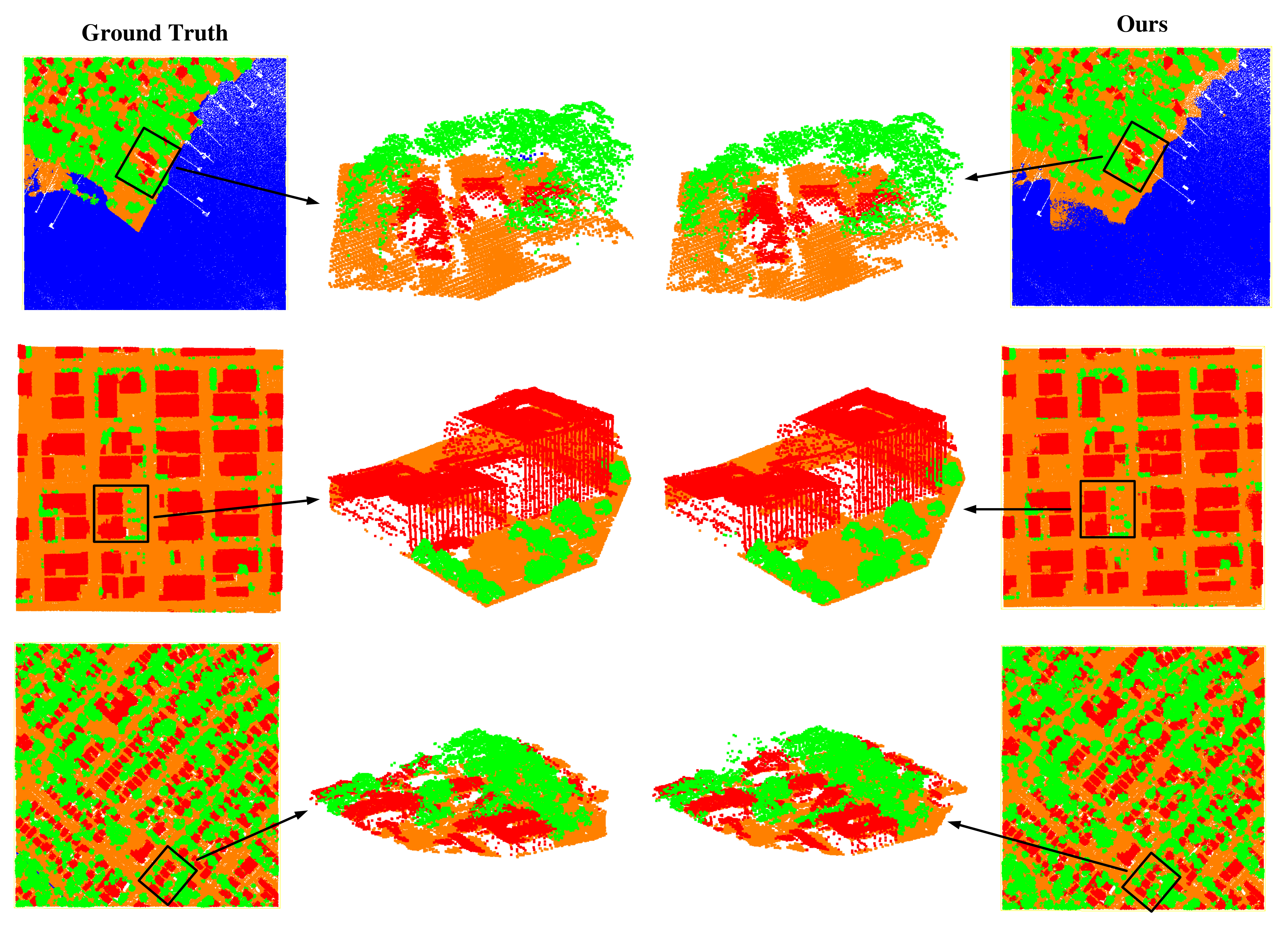}
  \caption{The classification results of our RFFS-Net on the DFC2019 dataset. The ground truth labels are shown in the first column, the local regions of the ground truth are shown in the second column, the local regions of the visualizations of our RFFS-Net are shown in the third column, and the visualizations of our RFFS-Net results are shown in the last column. From the visualizations, the classification results of our RFFS-Net are almost consistent with the ground truth.}
  \label{FIG:11}
\end{figure*}

\begin{table*}[htb]
  \caption{\textbf{Ablation studies of modules in the proposed approach.} "MRFALoss" denotes multi-level receptive field aggregation loss, "DAGFusion w/o dense" denotes dilated and annular graph fusion without dense connection, and "DAGFusion w/ dense" denotes dilated and annular graph fusion with dense connection.}\label{tbl5}
  \centering
  \begin{tabular}{ccc|ccc|ccc}
      \hline
      MRFALoss &DAGFusion w/o dense &DAGFusion w/ dense &mIoU &$\Delta$(mIoU) &$\Sigma\Delta$(mIoU) &mF1 & $\Delta$(mF1) &$\Sigma\Delta$(mF1) \\ 
      \hline
                                   &             &           & 52.8      &       &      & 66.3       &         &      \\ 
              $\surd  $            &             &           & 56.1      & +3.3  & 3.3  & 69.7       &+3.4     &3.4   \\ 
              $\surd  $            &$\surd  $    &           & 57.5      & +1.4  & 4.7  & 70.8       &+1.1     &4.5   \\          
              $\surd  $            &$\surd  $    & $\surd  $ & \bf{58.2} & +0.7  & 5.4  & \bf{71.6}  &+0.8     &5.3   \\ 
      \hline
  \end{tabular}
\end{table*}

\subsection{Results for the DFC2019 dataset}
As depicted in Table \ref{tbl4}, our RFFS-Net is compared with other classification networks (including PointNet++ ~\cite{qi2017pointnet++}, PointSIFT ~\cite{jiang2018pointsift}, PointCNN ~\cite{li2018pointcnn}, KPConv ~\cite{thomas2019kpconv}, DGCNN ~\cite{wang2019dynamic}, D-FCN ~\cite{wen2020directionally}, PointConv ~\cite{wu2019pointconv} and DANCE-NET ~\cite{li2020dance}) on the DFC2019 dataset. Through data statistics, we know that the number of category samples in the DFC2019 dataset is unbalanced, so we pay more attention to the mF1 and mIoU evaluation metric instead of the overall accuracy (OA). It is worth noting that our method achieves the best classification performance on mF1, reaching 88.04\% mF1 which outperforms PointConv by 1.94\%. In particular, our RFFS-Net reaches a high F1 value in the two categories of buildings and water while ensuring high performance in other categories. According to the characteristics of the data, water and ground are two categories with similar spatial structure (especially with respect to the height coordinates). Because our RFFS-Net can perceive a larger local region or even a global region, it can better distinguish such instances with similar spatial structures, which greatly reduces the misclassification of water. Besides, the superior performance of the building instances once again proves the superior distinguishing ability of our RFFS-Net for regions with complex structures.

The results achieved by our RFFS-Net are visualized in Fig. \ref{FIG:11}. After zooming in the local regions, we find that the prediction results of our method are almost consistent with the ground truth.

\section{Ablation study}
In this section, the effectiveness of our proposed multi-level receptive field aggregation loss (MRFALoss) and dilated and annular graph fusion module (DAGFusion) is first verified through experiments on ISPRS Vaihingen 3D. 
Then, we demonstrate the superiority of our RFFS-Net from the perspectives of visualization, the convergence of our model and model complexity.

\subsection{Effectiveness of the MRFALoss} 
In Table \ref{tbl5}, with MRFALoss, RFFS-Net improves the classification performance by 3.4\% (69.7\% vs. 66.3\%) on mF1 and 3.3\% (56.1\% vs. 52.8\%) on mIoU, which validates that minimizing the MRFALoss drives better multi-scale feature expression. The reason lies in the fact that the MRFALoss drives our RFFS-Net to give full play to the role of receptive field information. As mentioned in the previous analysis, our receptive field is calculated based on point sets of different resolutions, and point set labels of different scales are used as supervision information. This can leverage the receptive field of the deepest features towards points of different resolutions, and solve the problem of misclassification for instances of different scales.

\textbf{Weights of MRFALoss.}
In Table ~\ref{tbl6}, an ablation study is carried out to determine the weights of MRFALoss. Through the ablation study, we find the best combination of weights is \{1,0.3,0.3,0.3\}. Compared with other weights combinations, the weights combination \{1,0.3,0.3,0.3\} can make better use of supervised information at different resolutions, so it achieves better classification performance.

\begin{table}[htb]
  \small   
  \caption{\textbf{Ablation study on weights of MRFALoss.}}\label{tbl6}
  \centering
    \renewcommand{\arraystretch}{1.2}
    \begin{tabular}{c|cc}
    \hline
    Weights &mIoU &mF1\\  
    \hline
    $\lambda_1=1.0, \lambda_2=0.5, \lambda_3=1.0, \lambda_4=1.5$         &55.6      & 69.2     \\ 
    $\lambda_1=1.0, \lambda_2=1.5, \lambda_3=1.0, \lambda_4=0.5$         &56.6      & 70.1     \\ 
    $\lambda_1=1.0, \lambda_2=1.0, \lambda_3=1.0, \lambda_4=1.0$         &57.7      & 71.3     \\ 
    $\lambda_1=1.0, \lambda_2=1.5, \lambda_3=1.5, \lambda_4=1.5$         &55.8      & 69.4     \\ 
    $\lambda_1=1.0, \lambda_2=2.0, \lambda_3=2.0, \lambda_4=2.0$         &56.0      & 69.6     \\ 
    $\lambda_1=1.0, \lambda_2=0.5, \lambda_3=0.5, \lambda_4=0.5$         &57.9      & 71.4     \\ 
    $\bf{\lambda_1=1.0, \lambda_2=0.3, \lambda_3=0.3, \lambda_4=0.3}$    &\bf{58.2} & \bf{71.6}\\ 
    \hline
  \end{tabular}
\end{table}
\begin{figure}
  \centering
  \includegraphics[scale=.15]{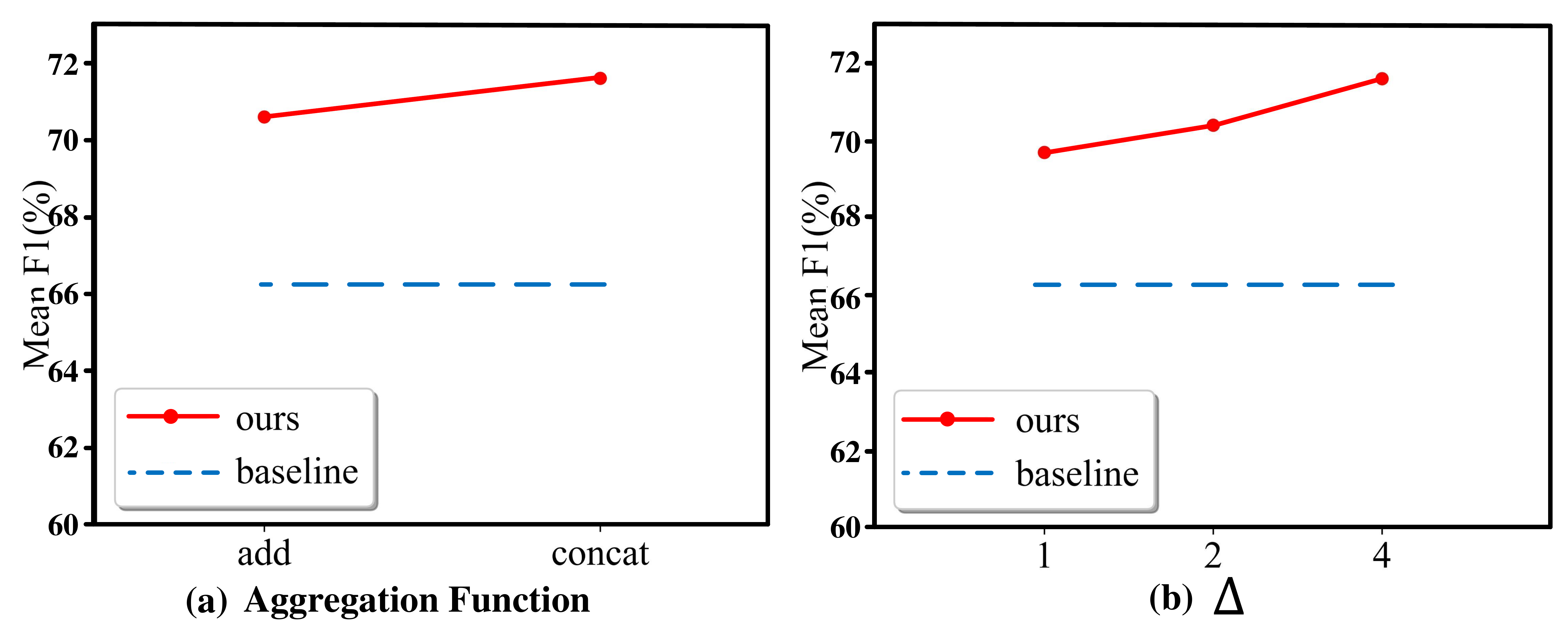}
  \caption{Evaluation of hyperparameters and modules. (a) Aggregation function. (b) Sampling step $\Delta$.}
\label{FIG:13}
\end{figure}
\begin{figure*}
	\centering
	  \includegraphics[width=0.9\textwidth]{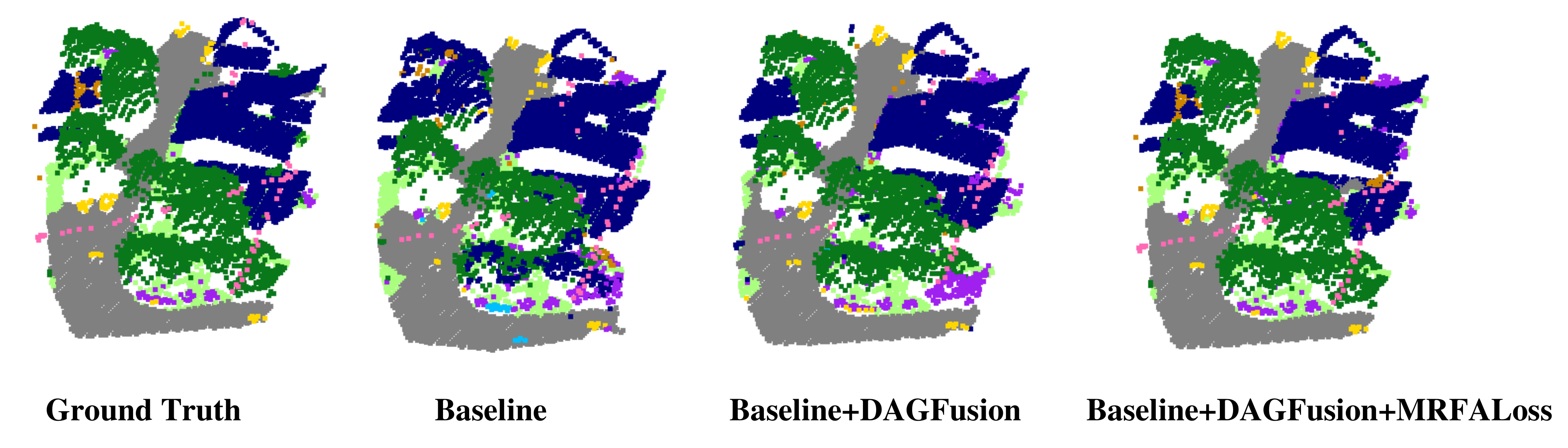}
	\caption{The visual comparison of the ground truth and the classification results achieved by the baseline, baseline+DAGFusion and baseline+DAGFusion+MRFALoss. The visualized results show that our DAGFusion can better adapt to areas with abundant structures. In addition, our MRFALoss can better adapt to instances with extreme scale variations (such as powerline with sparse points).}
	\label{FIG:12}
\end{figure*}

\begin{figure*}
	\centering
	  \includegraphics[width=1.0\textwidth]{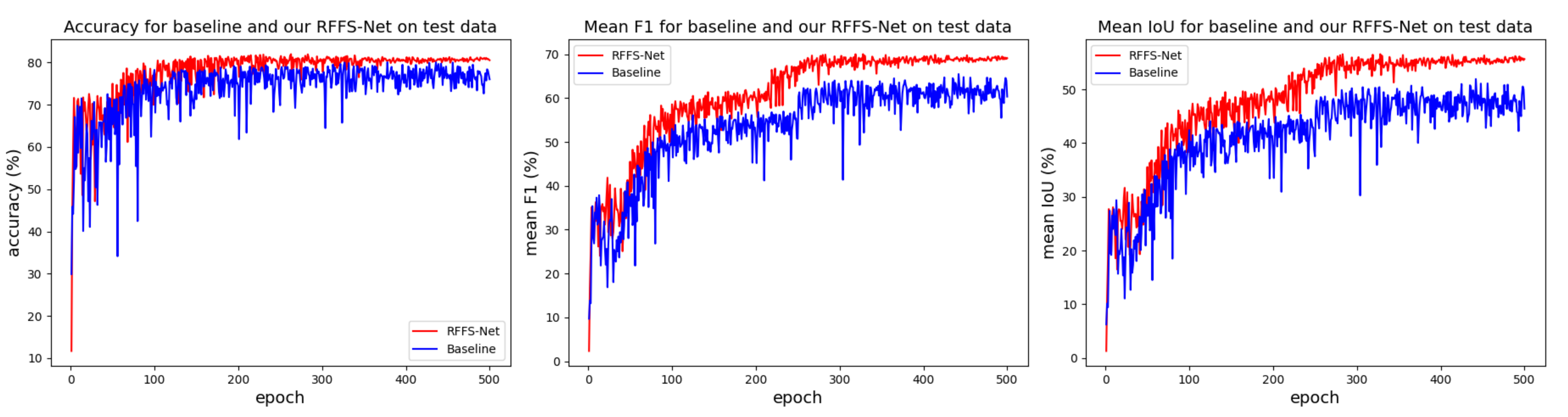}
	\caption{Convergence analysis between the baseline and our RFFS-Net on ISPRS Vaihingen 3D test set.}
	\label{FIG:14}
\end{figure*}
\subsection{Effectiveness of DAGFusion} 
From Table \ref{tbl5}, DAGFusion without dense connection improves the performance by 1.1\% (70.8\% vs. 69.7\%) on mF1 and 1.4\% (57.5\% vs. 56.1\%) on mIoU, which validates that the feature extraction performance of multi-scale receptive fields is better than that of using single receptive field. By adding the dense connection, DAGFusion with dense connection further improves the classification performance by 0.8\% (71.6\% vs. 70.8\%) on mF1 and 0.7\% (58.2\% vs. 57.5\%) on mIoU, which validates that dense connection brings denser feature pyramid and larger receptive field. In total, DAGFusion improves the performance by 1.9\% (71.6\% vs. 69.7\%) on mF1 and 2.1\% (58.2\% vs. 56.1\%) on mIoU, which are significant margins. This clearly demonstrates the superiority of the proposed DAGFusion over previous single receptive field methods. Besides, DAGFusion enhances the propagation of features and realizes the reuse of features. 

Before embedding DAGFusion, the network can only perceive a single local neighborhood, with a small single receptive field. 
However, after DAGFusion performs the fusion of different scales of receptive fields, the network can capture the features with different scales of receptive fields. Therefore, our network has superior performance for better distinguishing areas with complex structures and extreme scale variations.

\textbf{Sampling Step $\Delta$.}
In the construction of the DGConv dilated graphs, different step $\Delta$ values determine the different sparsity of the dilated graphs. Through experiments, we prove that, when the step $\Delta$ value is 4, the sparsity of the constructed dilated graphs can best meet the needs of the network and reach the best mF1 (71.6\%), as shown in Fig. \ref{FIG:13}(b). With the step $\Delta$ set to 4, the dilated graphs we obtain will not be excessively sparse. But the step cannot be too large, because a larger step will make DGConv tend to behave like a classical convolution, which will not achieve the purpose of expanding the receptive field.

\textbf{Aggregation function.}
As shown in Fig. \ref{FIG:13}(a), we experimentally validate that using the concatenate operation as the aggregate function of all DGConv or ADConv outputs can achieve the best performance. Compared with the add operation, the concatenate operation prevents the feature maps from becoming mixed up. Although the add operation achieves the effect of feature aggregation, it has a large confounding effect on the feature map.

\begin{figure*}
	\centering
	  \includegraphics[width=1.0\textwidth]{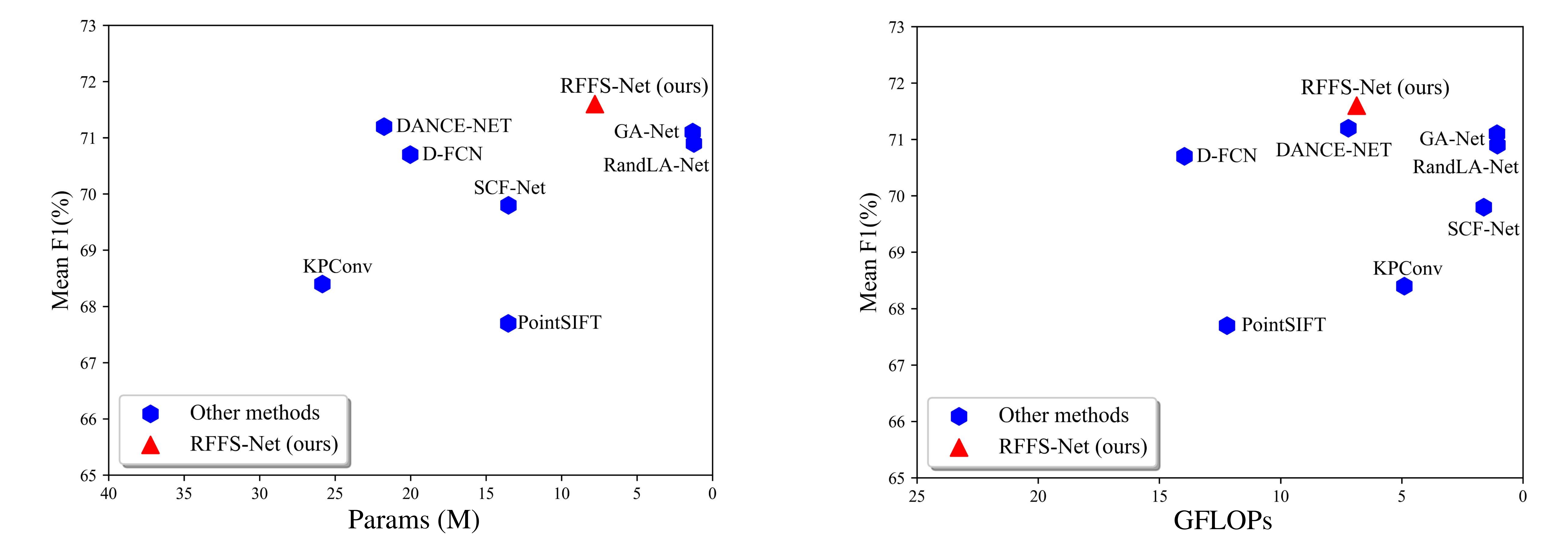}
	\caption{The number of parameters and FLOPs of different point classification methods. Blue hexagons represent other methods on ISPRS Vaihingen 3D. The red triangle is our RFFS-Net.}
	\label{FIG:15}
\end{figure*}

\begin{table}[htb]
  \small   
  \caption{\textbf{Ablation study with a focus on dilation combinations.} "Dilation Combinations" denotes the combinations of dilation rates.}
  \label{tbl7}
  \centering
    \renewcommand{\arraystretch}{1.2}
    \begin{tabular}{c|cc}
    \hline
    Dilation Combinations & mIoU &mF1 \\  
    \hline
    \{1,2\}          & 56.3       & 69.8\\ 
    \{2,4\}          & 56.2       & 69.9 \\ 
    \{1,4\}          & 57.6       & 71.3 \\ 
    \bf{\{1,2,4,8\}} & \bf{58.2}  & \bf{71.6} \\     
    \hline
  \end{tabular}
\end{table}

\textbf{Dilation combinations.}
In Table ~\ref{tbl7}, an ablation study is carried out to determine the dilation combinations of DAGFusion. Through the ablation study, we find the best combination of dilation rates is $\{1,2,4,8\}$. Compared with the dilation combinations \{1,2\},\{2,4\},\{1,4\}, the dilation combination \{1,2,4,8\} can capture receptive fields of more scales, so 
it achieves better classification performance.

\subsection{Visual analysis}
In Fig. \ref{FIG:12}, the visualizations of the ground truth as well as the classification results achieved by the baseline, baseline+DAGFusion, and baseline+DAGFusion+MRFALoss are given. Compared with the ground truth, our baseline has a poorer classification effect in areas with complex structures and extreme scale variations. By adding DAGFusion, the baseline with DAGFusion improves the classification performance, especially for areas with complex structures. Further, the baseline with DAGFusion and MRFALoss improves the classification performance for instances with extreme scale variations (such as powerline with sparse points).

\subsection{Convergence analysis}
We visualize the accuracy, mF1 and mIoU variation curves of the baseline and our RFFS-Net on ISPRS Vaihingen 3D test set. As shown in Fig. ~\ref{FIG:14}, during training, our RFFS-Net can consistently maintain the test performance higher than the baseline on OA, mF1 and mIoU. Not only that, from Fig. ~\ref{FIG:14}, we can also conclude that our RFFS-Net has more stable peaks and less glitches. This proves the better convergence and stability of our RFFS-Net.

\subsection{Network complexity}
RFFS-Net is a light classification network with three downsampling operations. As shown in Fig. \ref{FIG:15}, the existing classification methods introduce a lot of parameters and FLOPs. The extra parameters and FLOPs greatly limit their scope of application and practicality. In contrast, our RFFS-Net achieves the best performance while having significantly fewer parameters and FLOPs.

\section{Conclusion}
We propose RFFS-Net, which is an elegant and effective network to capture multi-scale receptive field features. By introducing the novel dilated graph convolution, RFFS-Net implements adaptive feature representation in accordance with dilated and annular graphs constructed by the Sparse-KNN search strategy. With dilated and annular graph fusion in the receptive field fusion process, RFFS-Net implements the aggregation of multi-receptive field features. By optimizing the proposed MRFALoss in the receptive field stratification process, RFFS-Net realizes the optimized expression of the receptive field information with each resolution point set as calculation basis. Extensive experiments on commonly used benchmarks (ISPRS Vaihingen 3D, LASDU, DFC2019) validated RFFS-Net's superior performance. This is in striking contrast with the state-of-the-art 4-downsampling-layers classification networks. RFFS-Net provides a fresh insight for the expression of multi-receptive field features.

In the future, we hope to design more efficient network architectures, such as taking into account the uneven density of point clouds in local feature learning. In addition, we also need to take into account the large variations in elevation of point clouds that our method does not focus on, and hope to propose an elevation-aware method to reduce the problem of misclassification caused by large variations in vertical heights.

\section{Acknowledgement}
This work for airborne laser scanning point cloud classification was supported by the National Natural Science Foundation of China (NSFC) under Grant 62171436.

\bibliographystyle{IEEEtran}
\bibliography{egbib.bbl}

\end{document}